\documentclass{article}
\usepackage{spconf,amsmath,graphicx}

\usepackage{amsmath,amssymb,amsfonts}
\usepackage{algorithmic}
\usepackage{algorithm}
\usepackage{cite}

\usepackage[dvipsnames]{xcolor}
\usepackage{pifont}  % For check/cross symbols

\newcommand{\cmark}{\textcolor{ForestGreen}{\ding{51}}}
\newcommand{\xmark}{\textcolor{Red}{\ding{55}}}

\usepackage{booktabs}
\usepackage{nicefrac}
\usepackage{microtype}
\usepackage{array,multirow,colortbl,arydshln}
\usepackage[export]{adjustbox}
\usepackage{wrapfig}
\usepackage{enumitem}
\usepackage{bbm}
\usepackage{caption}
\usepackage{hyperref}
\usepackage{url}

\usepackage{geometry}
\title{COT-AD: Cotton  Analysis Dataset}

\name{
\begin{tabular}{c}
Akbar Ali$^{\star}$ \quad Mahek Vyas$^{\wedge}$ \quad Soumyaratna Debnath$^{\star}$ \quad Chanda Grover Kamra$^{\dagger}$ \\
Jaidev Sanjay Khalane$^{\star}$ \quad Reuben Shibu Devanesan$^{\star}$ \quad Indra Deep Mastan$^{\ddagger}$ \\
Subramanian Sankaranarayanan$^{\star}$ \quad Pankaj Khanna$^{\star}$ \quad Shanmuganathan Raman$^{\star}$
\end{tabular}
}

\address{
\begin{tabular}{c}
$^{\star}$ Indian Institute of Technology Gandhinagar, India. \quad
$^{\dagger}$ Ashoka University, Sonipat, India. \\
$^{\wedge}$ Larsen \& Toubro Technology Services Limited. \quad
$^{\ddagger}$ Indian Institute of Technology (BHU), Varanasi, India.
\end{tabular}
}

\begin{document}
%\ninept
%
\maketitle

\begin{abstract}

This paper presents COT-AD, a comprehensive Dataset designed to enhance cotton crop analysis through computer vision. Comprising over 25,000 images captured throughout the cotton growth cycle, with 5,000 annotated images, COT-AD includes aerial imagery for field-scale detection and segmentation and high-resolution DSLR images documenting key diseases. The annotations cover pest and disease recognition, vegetation, and weed analysis, addressing a critical gap in cotton-specific agricultural datasets. COT-AD supports tasks such as classification, segmentation, image restoration, enhancement, deep generative model-based cotton crop synthesis, and early disease management, advancing data-driven crop management
.The COT-AD dataset can be found here:\url{https://aamaanakbar.github.io/COT-AD/}.

\end{abstract}
\section{Introduction}
\label{sec:intro}

Cotton is a critical global crop, essential for the textile industry and the economies of many countries like India, China, and Brazil. It supports over 250 million people worldwide, including farmers and processors. However, cotton farming\cite{li2022yield} faces numerous challenges, such as pest infestations, diseases, and climate change. 
% For example, uncontrolled weed growth\cite{dang2023yoloweeds} can reduce yields by 50-85\% while biotic stress can cause up to a 30\%  Despite a modest increase in global production forecasts, the sector is threatened by environmental sustainability issues that could affect its long-term viability.

\begin{table*}[!ht]
\centering
\renewcommand{\arraystretch}{1.3} % Adjust row height for better readability
\setlength{\tabcolsep}{8pt} % Adjust column spacing for better readability
\resizebox{\linewidth}{!}{%
\begin{tabular}{llcccccccc}
\hline
\multicolumn{2}{c}{\multirow{2}{*}{\textbf{Dataset}}} & \multirow{2}{*}{\textbf{Images}} & \multirow{2}{*}{\textbf{Classes}} & \multirow{2}{*}{\textbf{Type}} & \multicolumn{5}{c}{\textbf{Tasks}} \\ \cline{6-10} 
\multicolumn{2}{c}{} &  &  &  & \textbf{Classification} & \textbf{Detection} & \textbf{Segmentation} & \textbf{Enhancement} & \textbf{Synthesis} \\ \hline
\multicolumn{2}{l}{Bishshash et al.\cite{bishshash2024comprehensive }} & 2137 & 8 & Hand Held & \color{ForestGreen}{\cmark} & \color{Red}{\xmark} & \color{Red}{\xmark} & \color{Red}{\xmark} & \color{Red}{\xmark} \\
\multicolumn{2}{l}{Mirani et al.\cite{miranidataset}} & 4000 & 7 & Hand Held & \color{ForestGreen}{\cmark} & \color{Red}{\xmark} & \color{Red}{\xmark} & \color{Red}{\xmark} & \color{Red}{\xmark} \\
% \multicolumn{2}{l}{Ali et al. (Kaggle)\textcolor{red}{cite??}} & 1345 & 12 & Hand Held & \color{ForestGreen}{\cmark} & \color{Red}{\xmark} & \color{Red}{\xmark} & \color{Red}{\xmark} & \color{Red}{\xmark} \\
% \multicolumn{2}{l}{Roboflow (college-kcu84)} & 982 & 1 & Hand Held & \color{Red}{\xmark} & \color{ForestGreen}{\cmark} & \color{Red}{\xmark} & \color{Red}{\xmark} & \color{Red}{\xmark} \\
\multicolumn{2}{l}{Li et al.\cite{li2022yield}} & 4000 & 1 & Aerial & \color{Red}{\xmark} & \color{Red}{\xmark} & \color{ForestGreen}{\cmark} & \color{Red}{\xmark} & \color{Red}{\xmark}\\
\multicolumn{2}{l}{Karim et al.\cite{cotton_leaf_disease_dataset}} & 1710 & 4 & Hand Held & \color{ForestGreen}{\cmark} & \color{Red}{\xmark} & \color{Red}{\xmark} & \color{Red}{\xmark} & \color{Red}{\xmark}\\
% \multicolumn{2}{l}{PlantSeg Datasets \textcolor{red}{cite??}} & 110 & 2 & Hand Held & \color{Red}{\xmark} & \color{Red}{\xmark} & \color{ForestGreen}{\cmark} & \color{Red}{\xmark} & \color{Red}{\xmark} \\
\multicolumn{2}{l}{\textbf{COT-AD(Ours)}} & 25000+ & {9} & Aerial + Hand Held & \color{ForestGreen}{\cmark} & \color{ForestGreen}{\cmark} & \color{ForestGreen}{\cmark} & \color{ForestGreen}{\cmark} & \color{ForestGreen}{\cmark} \\ \hline
\end{tabular}%
}
\caption{Specifications of available datasets for various tasks, highlighting the COT-AD dataset with over 25,000 images and comprehensive support for classification, detection, segmentation, image enhancement, and cotton disease detection.}
\label{table:datasets1}
\end{table*}

\begin{table*}[!ht]
\centering
\renewcommand{\arraystretch}{1.3} % Adjust row height for better readability
\setlength{\tabcolsep}{8pt} % Adjust column spacing for better readability
\resizebox{\linewidth}{!}{%
\begin{tabular}{llcccccccc}
\hline
\multicolumn{2}{c}{\multirow{2}{*}{\textbf{Split}}} & \multirow{2}{*}{\textbf{Images}} & \multirow{2}{*}{\textbf{Classes}} & \multirow{2}{*}{\textbf{Type}} & \multicolumn{5}{c}{\textbf{Tasks}} \\ \cline{6-10} 
\multicolumn{2}{c}{} &  &  &  & \textbf{Classification} & \textbf{Detection} & \textbf{Segmentation} & \textbf{Enhancement} & \textbf{Synthesis} \\ \hline
\multicolumn{2}{l}{Drone} & 1800 & 1 & Aerial & \color{Red}{\xmark} & \color{ForestGreen}{\cmark} & \color{ForestGreen}{\cmark} & \color{Red}{\xmark} & \color{Red}{\xmark} \\
\multicolumn{2}{l}{DSLR Disease} & 3231 & 9 & Hand Held & \color{ForestGreen}{\cmark} & \color{ForestGreen}{\cmark} & \color{Red}{\xmark} & \color{Red}{\xmark} & \color{Red}{\xmark}\\
\multicolumn{2}{l}{DSLR Segment} & 100 & 12 & Hand Held & \color{Red}{\xmark} & \color{Red}{\xmark} & \color{ForestGreen}{\cmark} & \color{Red}{\xmark} & \color{Red}{\xmark}\\
\multicolumn{2}{l}{Drone + DSLR} & 14000+11000 & - & Aerial + Hand Held & \color{Red}{\xmark} & \color{Red}{\xmark} & \color{Red}{\xmark} & \color{ForestGreen}{\cmark} & \color{ForestGreen}{\cmark} \\ \hline
\end{tabular}%
}
\caption{Specifications of the COT-AD dataset splits, including Drone, DSLR Disease, DSLR Segment, and Synthesis, with varied image counts and task support for detection, segmentation, and conditional synthesis, showcasing the dataset's versatility across different imaging types and tasks.}
\label{table:datasets2}
\end{table*}

Smart farming technologies help manage these challenges by improving precision in nutrient application, monitoring plant stress, and predicting yields. Tools like soil sensors, drones with RGB and multispectral cameras, and satellite imagery allow \cite{li2022yield} for targeted interventions, such as irrigation or pesticide application, and more reliable yield forecasts. Precision farming methods can be categorized as intrusive (e.g., manual soil sampling) and non-intrusive (e.g., remote sensing via drones), with the latter offering lower environmental impact and cost.

AI-driven precision farming optimizes practices \cite{dang2023yoloweeds} through machine learning and data analytics. This includes crop prediction, pest detection, precision irrigation, and autonomous machinery for planting and harvesting. However, existing datasets(Table~\ref{table:datasets1})  for cotton image analysis are limited in scope and quality. Current datasets primarily focus on single tasks, with \textbf{``inadequate volume of data''} and limited resolution, hindering broader applications like disease detection, cotton segmentation, species classification, synthetic image generation, and text-guided enhancement. More diverse datasets are needed to support these advanced tasks.

\begin{figure}[t]
  \includegraphics[width=\linewidth]{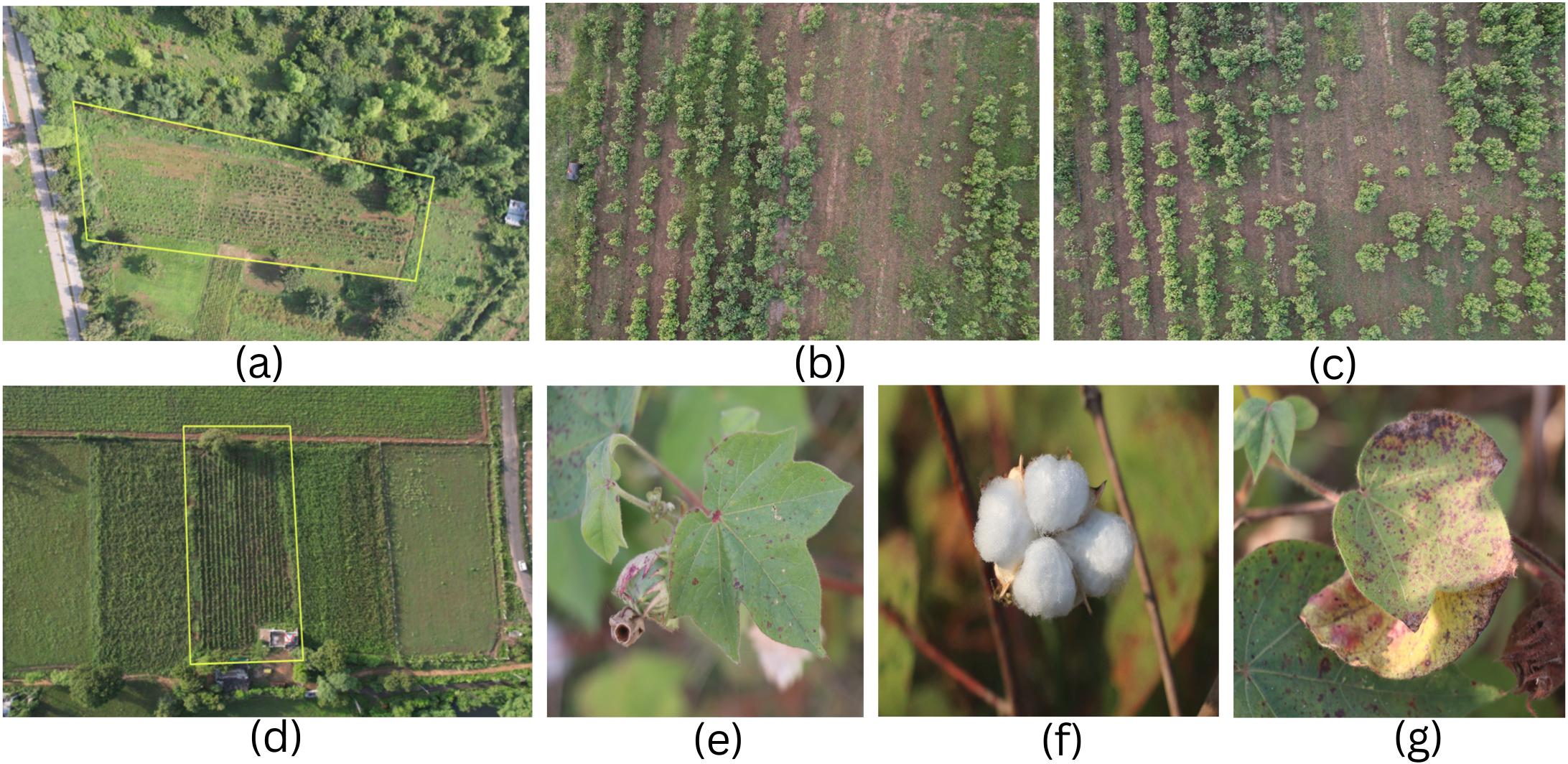}
  \caption{Part \textbf{(a)} showcases the field view of Site 1; Parts \textbf{(b)} and \textbf{(c)} are examples of drone-captured images; Part \textbf{(d)} showcases the field view of Site 2; and Parts \textbf{(e)}, \textbf{(f)}, and \textbf{(g)} are examples of DSLR images.}
  \label{fig:site_and_data}
\end{figure}

This paper introduces the COT-AD dataset\footnote{The dataset is available for non-commercial use on both \href{https://ieee-dataport.org/documents/cot-adcotton-analysis-dataset}{IEEE DataPort} and \href{https://www.kaggle.com/datasets/aamaanakbar/cot-ad1}{Kaggle}.}
 for cotton crop detection \cite{dang2023yoloweeds}, segmentation, image restoration, enhancement and disease classification. Our proposed dataset enables machines to envision the visual and semantic information about crops. The dataset includes aerial images (See Table~\ref{tab:aerial_analysis}) of cotton fields captured throughout the crop's life cycle and close-up images(See Fig.~\ref{fig:site_and_data}) taken with a Digital Single-Lens Reflex (DSLR)  camera,   documenting various diseases that affect cotton crops. The major contributions are described as follows.

\begin{itemize}[leftmargin=*]
    
    \item We introduce a novel, large-scale dataset, COT-AD, featuring over 25,000 images collected from UAVs (aerial views) and DSLR cameras (close-up inspection) across the entire cotton growth cycle (Table~\ref{table:datasets2}).
    \vspace{-.2cm}
    \item We provide annotations in the COT-AD dataset for disease classification, cotton crop detection, and segmentation (Sec.~\ref{sec:COT-ADdataset}).
    \vspace{-.2cm}
    \item We perform diverse computer vision tasks such as crop detection, disease classification, segmentation, restoration, enhancement, and synthesis (Sec.~\ref{sec:applications}) on the COT-AD dataset.
\end{itemize}
    \vspace{-.35cm}
\section{Related Works}
%Cotton crop diseases significantly impact yield and quality, making their study essential due to the economic importance of cotton. 
Several key datasets support disease classification~\cite{li2022yield, cotton_leaf_disease_dataset}, including the Cotton Crop Plant Leaves Dataset\cite{li2022yield}, the Cotton Leaf Dataset\cite{miranidataset}, and the Cotton Disease Dataset\cite{bishshash2024comprehensive}. While the first two focus on leaf images, the third provides a broader scope by including both leaf and whole-plant images. Crop detection and segmentation\cite{dang2023yoloweeds} play a vital role in precision agriculture and plant phenotyping. Detection involves identifying and localizing plants, such as crops and weeds, facilitating health monitoring and automation, whereas segmentation differentiates plants from the background \cite{dang2023yoloweeds} for detailed analysis.

\section{COT-AD Dataset}
\label{sec:COT-ADdataset}
\begin{figure}[t]
  \includegraphics[width=\linewidth]{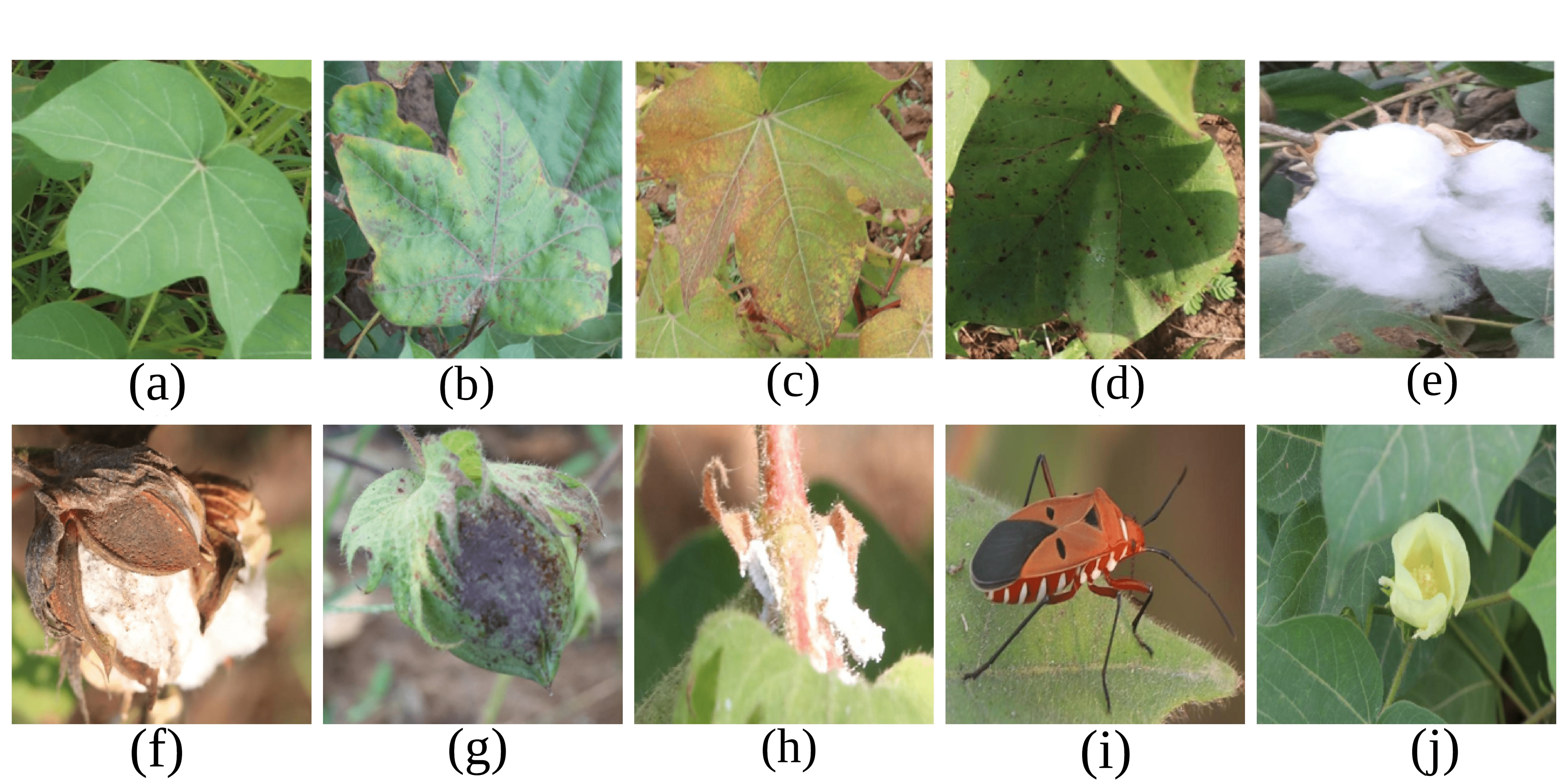}
  \caption{The figure shows example Images for Cotton Disease Classification: (a) Fresh Leaf, (b) Yellowish Leaf, (c) Leaf Reddening, (d) Leaf Spot/Bacterial Blight, (e) Healthy Cotton Boll, (f) Cotton Boll Rot , (g) Damaged Cotton Boll, (h) Mealy Bug, (i)Red Cotton Bug , and (j) Cotton Flower. }
  \label{fig:DSLR_images_sample_groups}
\end{figure}

The COT-AD dataset, shown in Table~\ref{table:datasets2}, significantly expands upon previous datasets (outlined in Table~\ref{table:datasets1}) by offering over $25,000$ images captured using drone and DSLR cameras. It includes high-resolution aerial imagery collected at altitudes of [$10$m, $15$m, and $115$m] and detailed close-up images from handheld DSLR cameras. Furthermore, the field videos in the dataset help perform dynamic analysis that includes disease detection, crop health determination, classification and segmentation, restoration, and even synthesis.
Compared to previous datasets Table~\ref{table:datasets1}, COT-AD is more prosperous and agile and can be utilized for different applications in cotton science and agrotech. Such a combination of aerial and close-range photos and the larger volume makes it favorable for machine learning tasks and AI-enabled technologies, making it accessible for researchers, practitioners, and developers of agricultural AI.\\~
\vspace{-.01cm}
\noindent \textbf{Data Acquisition, Annotation.}    The COT-AD dataset was collected using a DJI Mavic Air 2 drone over six months in two cotton fields: Organic Farm (weekly data) and Kiran Farm (bi-weekly data). We captured data at varying altitudes to assess altitude effects. Close-up images of cotton crops were taken twice a week on the Organic Farm using a Canon EOS 80D DSLR, with experts validating and categorizing the images. The dataset includes images representing different aspects of cotton crop health, aiding growth monitoring and issue detection. The COT-AD dataset images are categorized into four main groups, namely Cotton Leaf, Cotton Boll, Cotton Flower, and Bugs (Table~\ref{tab:cotton_categories} and Fig.~\ref{fig:DSLR_images_sample_groups}) for monitoring cotton health.

Once the imagery was collected, the annotation process began. Each image was manually annotated, labelling the cotton crops within the images. Manual annotation ensures high data accuracy, crucial for training effective Deep learning models. 

The folder structure for the annotated data is organized under the \textit{Detection and Segmentation Task} directory, which is divided into four main parts: Part A (data from the first two months), Part B (data from the third month), Part C (data from the fourth month), and Part D (data from the fifth and sixth months). A comprehensive specification of the dataset structure is included in the supplementary materials. 
%% \begin{itemize}
%%     \item \textbf{Cotton Leaf:} \textit{Yellow-ish Leaf}, \textit{Bacterial Blight/Leaf Spot}, \textit{Leaf Reddening}, \textit{Fresh Leaf}.    
%%     \item \textbf{Cotton Boll:} \textit{Boll Rot}, \textit{Damaged Cotton Boll}, \textit{Healthy Cotton Boll}.   
%%     \item\textbf{Cotton Flower:} \textit{White Cotton Flower}, \textit{Red Cotton Flower}.  
%%     \item\textbf{Bugs:} \textit{Mealy Bug}, \textit{Red Cotton Bug}.
%% \end{itemize}

\begin{table}[!ht]
\centering
\resizebox{\columnwidth}{!}{%
\begin{tabular}{cccc}
\hline
\textbf{\begin{tabular}[c]{@{}c@{}}Age of Crop\\ (Months)\end{tabular}} & \textbf{\begin{tabular}[c]{@{}c@{}}Average Area \\ Covered by Crops \\ (per image)\end{tabular}} & \textbf{\begin{tabular}[c]{@{}c@{}}Average Number \\ of Crops \\ (per image)\end{tabular}} & \textbf{\begin{tabular}[c]{@{}c@{}}Total Instances \\ of Crops\end{tabular}} \\ \hline
Month 1 and 2 & 56.95\% & 49 & 14538 \\
Month 3 & 58.01\% & 58 & 37881 \\
Month 4 & 63.24\% & 73 & 27325 \\
Month 5 and 6 & 73.01\% & 44 & 21656 \\ \hline
\end{tabular}%
}
\caption{This table presents an analysis of \textbf{ Aerial Images} of the COT-AD dataset over different age intervals.}
\label{tab:aerial_analysis}
\end{table}

\begin{table}[!ht]
\renewcommand{\arraystretch}{1.1} % Adjust row height for better readability
    \centering
    \resizebox{\columnwidth}{!}{%
    \begin{tabular}{p{55pt} p{185pt}}
        \hline
        \textbf{Category} & \textbf{Types} \\
        \hline
        \textbf{Cotton Leaf} & Yellow-ish Leaf, Bacterial Blight/Leaf Spot, Leaf Reddening, Fresh Leaf \\
        \hline
        \textbf{Cotton Boll} & Boll Rot, Damaged Cotton Boll, Healthy Cotton Boll \\
        \hline
        \textbf{Cotton Flower} & White Cotton Flower, Red Cotton Flower \\
        \hline
        \textbf{Bugs} & Mealy Bug, Red Cotton Bug \\
        \hline
    \end{tabular}}
    \caption{Categorization of Cotton Plant Parts.}
    \label{tab:cotton_categories}
\end{table}

\begin{figure*}[!h]
  \includegraphics[width=\linewidth]{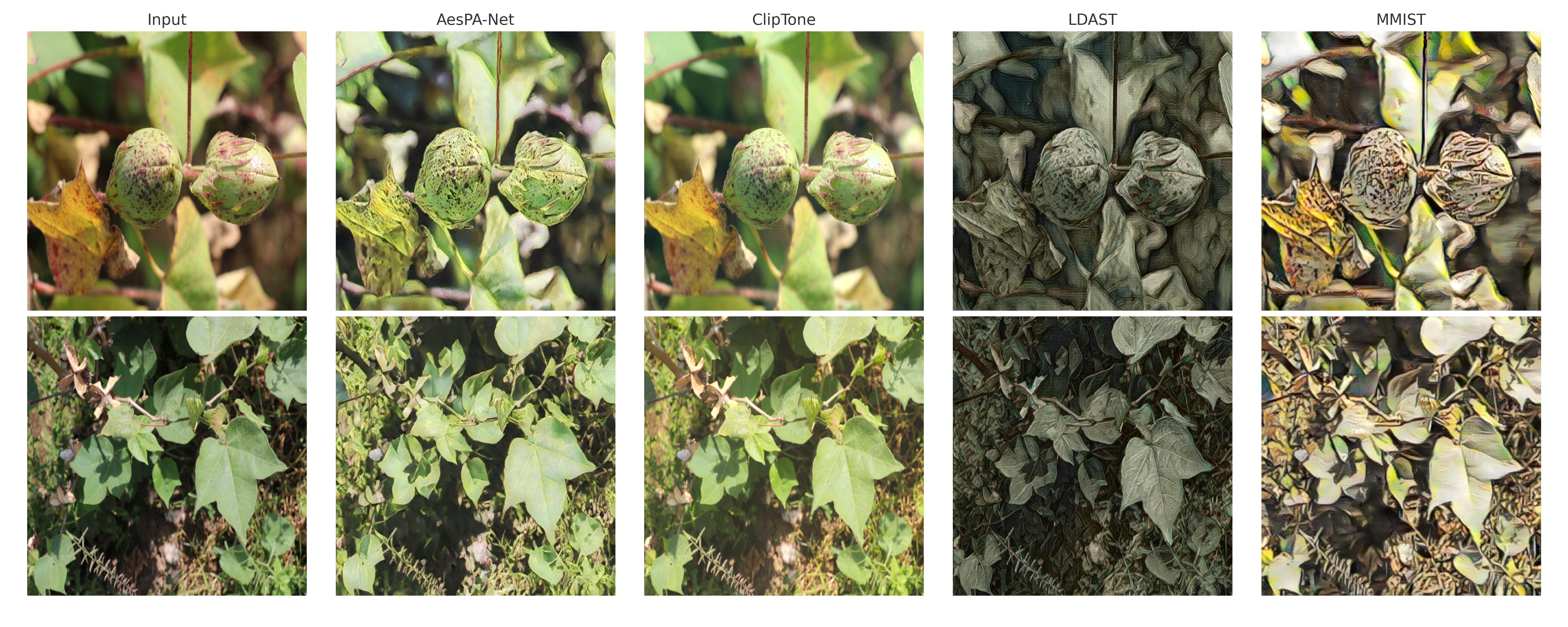}
  \caption{\textbf{Image Enhancement.} Each row has different text prompts for enhancement. Row 1: Generate a high-resolution image with sharp details, smooth textures, balanced contrast, proper lighting, and no noise or distortion. Row 2:  A clear version of the input image with improved visibility, contrast, and detail, effectively removing haze.}
    % Row-3: A clean, high-quality image with balanced contrast and brightness, free from noise or distortion.
    \label{fig:img_enh}   
\end{figure*}

\vspace{-.2cm}
\section{Applications}
\vspace{-.2cm}
This section presents the applications performed on the COT-AD dataset. We also provide additional results in the supplementary files. 
\label{sec:applications}
\vspace{-.2cm}
\subsection{Image Enhancement}
\vspace{-.1cm}
We enhance the quality of DSLR-collected cotton-crop images using a text-guided image stylization method. Existing methods such as mmist~\cite{wang2024multimodality}, and LDAST~\cite{ldast_ECCV2022} produce disharmonious artifacts due to low semantic correspondence between content and style images, as shown in Fig.~\ref{fig:img_enh}. To address this, we apply Aesthetic Pattern-Aware style transfer Networks (AesPA-Net)\cite{hong2023aespa}, which balances local and global style expressions, producing better results. We applied CLIPtone~\cite{Lee_2024_CVPR} to transfer the exact color tone of the text prompt as shown in column (e) of Fig.~\ref{fig:img_enh}. As seen in Table~\ref{table:imEnhance}, AesPA-Net outperforms other methods in visual quality and quantitative scores such as Arniqa~\cite{agnolucci2024arniqa}, Brisque~\cite{mittal2011blind}, and Clipiqa~\cite{wang2023exploring} scores. However, Clipscores~\cite{hessel2021clipscore} are higher for other methods due to their direct text prompt-based style features.

\begin{figure}[!ht]
  \includegraphics[width=\linewidth]{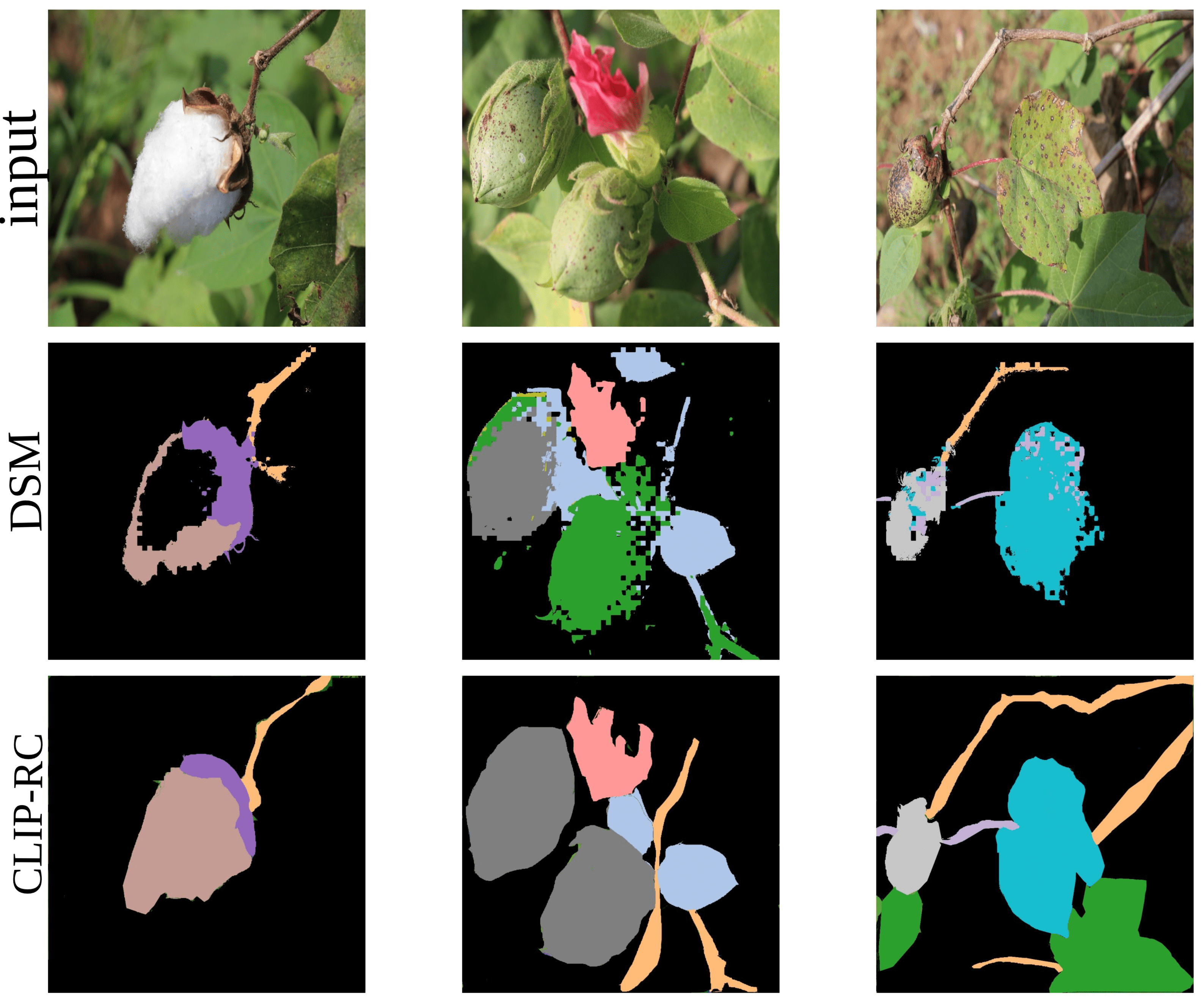}
  \caption{Semantic Segmentation Masks with Deep Spectral Method (DSM)~\cite{melas2022deep}. Each different color represents a semantic class of the dataset.}
\label{fig:semSeg_DSM}
\end{figure}

\begin{table}[!h]
\centering
\resizebox{0.49\textwidth}{!}{ % Adjust width to fit half-page
\begin{tabular}{p{70pt}p{30pt}p{30pt}p{30pt}p{30pt}p{35pt}}
\hline
\rowcolor[HTML]{FFFFFF} 
Method         & \textbf{Arniqa} \cite{agnolucci2024arniqa} & \textbf{Brisque} \cite{mittal2011blind} & \textbf{Clipiqa} \cite{wang2023exploring} & \textbf{Nima} \cite{talebi2018nima} & \textbf{Clipscore} \cite{hessel2021clipscore} \\ \hline
% Color-Transfer & 0.55            & 24.88            & 0.54             & 4.03          & 0.58               \\
MMIST \cite{wang2024multimodality}          & 0.59            & 22.5             & 0.60             & \textbf{4.20} & \textbf{0.64}      \\
LDAST \cite{ldast_ECCV2022}          & 0.52            & 34.78            & 0.51             & 5.21          & \textbf{0.64}      \\
CLIPStyler \cite{kwon2022clipstyler}     & 0.54            & 25.73            & 0.59             & 4.76          & 0.60             \\
% GT             & 0.55            & 29.05            & \textbf{0.69}    & 4.06          & 0.56               \\
AesPA-Net \cite{hong2023aespa}      & \textbf{0.65}   & \textbf{9.61}    & \textbf{0.65}             & 4.15          & 0.59               \\ \hline
\end{tabular}
}
\caption{Image Quality Assessment (IQA) scores for the Image Enhancement.}
\label{table:imEnhance}
\end{table}

\vspace{-.2cm}
\subsection{Image Segmentation}
\vspace{-.1cm}
\begin{figure}[t]
  \includegraphics[width=\linewidth]{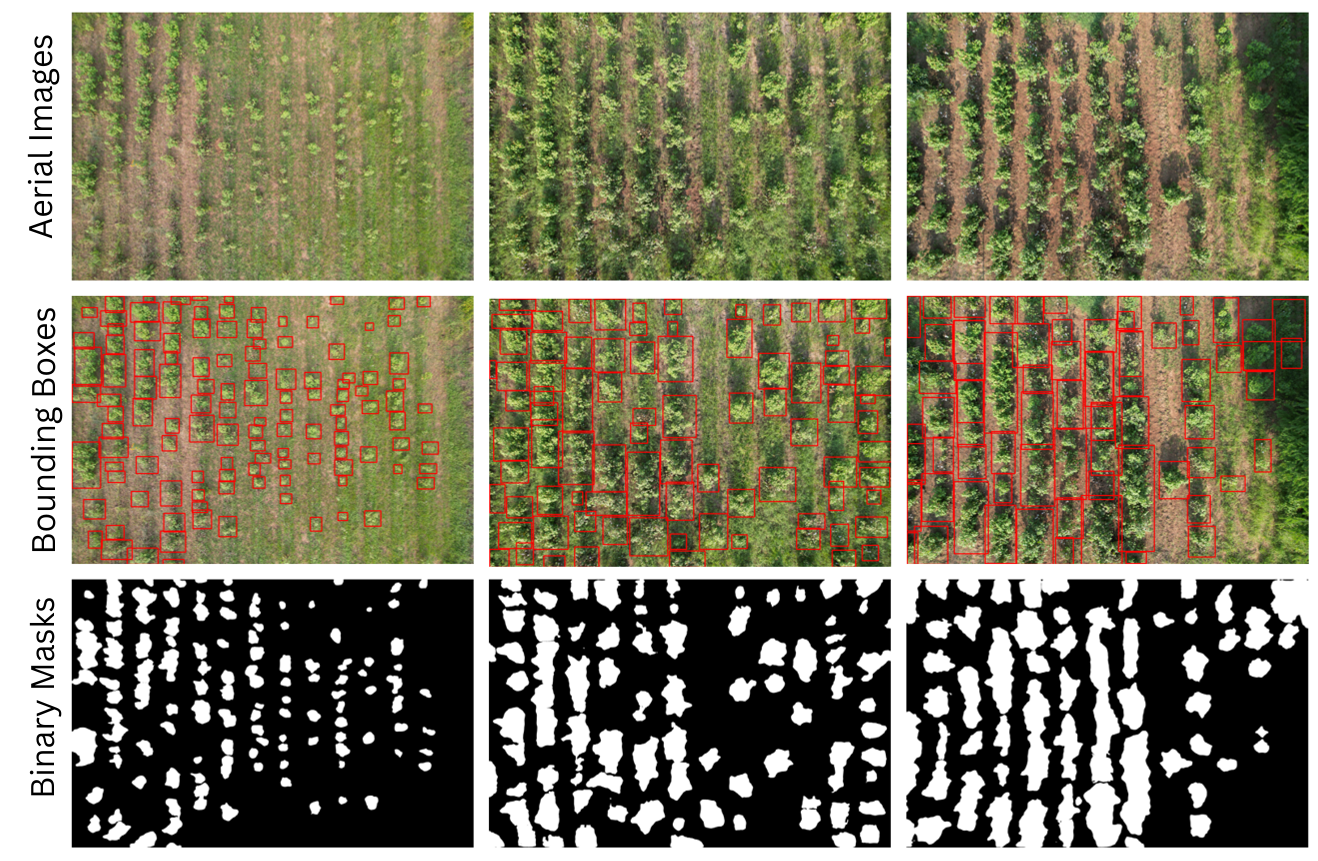}
  \caption{Overview of Detection and Segmentation Dataset - original aerial images of cotton crops, the corresponding detection bounding boxes, and the segmentation masks.}
  \label{fig:bb_mask_overview}
\end{figure}
% \begin{figure}
%   \includegraphics[width=\linewidth]{images/img_seg_res.png}
%   \caption{Comparison of segmentation results for YOLOv11 against the ground truth. The top row displays the ground truth annotations, and the bottom row illustrates predictions by YOLOv11. }
%   \label{fig:seg_performance}
% \end{figure}
We perform semantic segmentation in unimodal and multimodal settings. In unimodal, we use a strong unsupervised method for segmentation named Deep Spectral Method~\cite{melas2022deep}, and in multimodal settings, we have used CLIP-RC\cite{Zhang_2024_CVPR} used YOLOv11~\cite{dang2023yoloweeds}, as given in Fig.~\ref{fig:bb_mask_overview}. In the unsupervised setting, we applied the Deep Spectral method\cite{melas2022deep}, which extracts features from a self-supervised pre-trained network and uses spectral graph theory on feature correlations to obtain eigenvectors. These edge segments correspond to semantically meaningful regions with well-defined boundaries. We found that CLIP-RC\cite{Zhang_2024_CVPR} is able to give correct semantic classes in an image as compared to DSM~\cite{melas2022deep} as shown in Fig.~\ref{fig:semSeg_DSM}. 

\vspace{-.2cm}
\subsection{Image Classification} 
\vspace{-.2cm}
\begin{table}[!h]
\centering
\resizebox{0.49\textwidth}{!}{ % Adjust width to fit half-page
\begin{tabular}{ c c c }
\hline
\textbf{Model} & \textbf{\begin{tabular}[c]{@{}l@{}}Zero Shot\\  (Top-5 )\end{tabular}} & \textbf{\begin{tabular}[c]{@{}l@{}}Linear Probing\\ (Accuracy)\end{tabular}} \\
\hline
CLIP (ViT-32)~\cite{radford2021learning} & 93.99 & 89.79 \\
CLIP (ViT-16)~\cite{radford2021learning} & 95.88 & 92.27 \\
CLIP (RN50)~\cite{radford2021learning} & \textbf{97.61} & 86.39 \\
Imageomics/bioclip~\cite{stevens2024bioclip} & 65.27 & \textbf{91.96} \\
Bioclip-vit-b-16-inat-only~\cite{stevens2024bioclip} & 67.65 & 91.49 \\
\hline
\end{tabular}}
\caption{Model Performance: Zero-Shot and Linear Probing}
\label{tab:zeroshot_linearProbing}
\end{table}

\vspace{-.2cm}
\noindent \textbf{COT-AD Classification.} We conducted zero-shot classification and linear probing experiments to assess the generalization and representation capabilities of CLIP ~\cite{radford2021learning} and BioCLIP~\cite{stevens2024bioclip} models on the cotton-crop dataset. As shown in Table~\ref{tab:zeroshot_linearProbing}, , CLIP (RN50) achieves better top-5 accuracy than CLIP (ViT-16 or 32) for zero-shot experiment. This is because we are using only 100 image points to train the classifier, and CLIP (ViT) is getting overfitted, which uses more parameters than RN50. CLIP achieves higher top-5 accuracy, indicating better generalization, while BioCLIP outperforms CLIP in linear probing accuracy (column 3 of Table~\ref{tab:zeroshot_linearProbing}), suggesting a better representation of the cotton-crop dataset. This is expected since CLIP is trained on general image-text pairs, whereas BioCLIP is trained on the Trees of Life dataset.

\begin{figure}[t]
  \includegraphics[width=\linewidth]{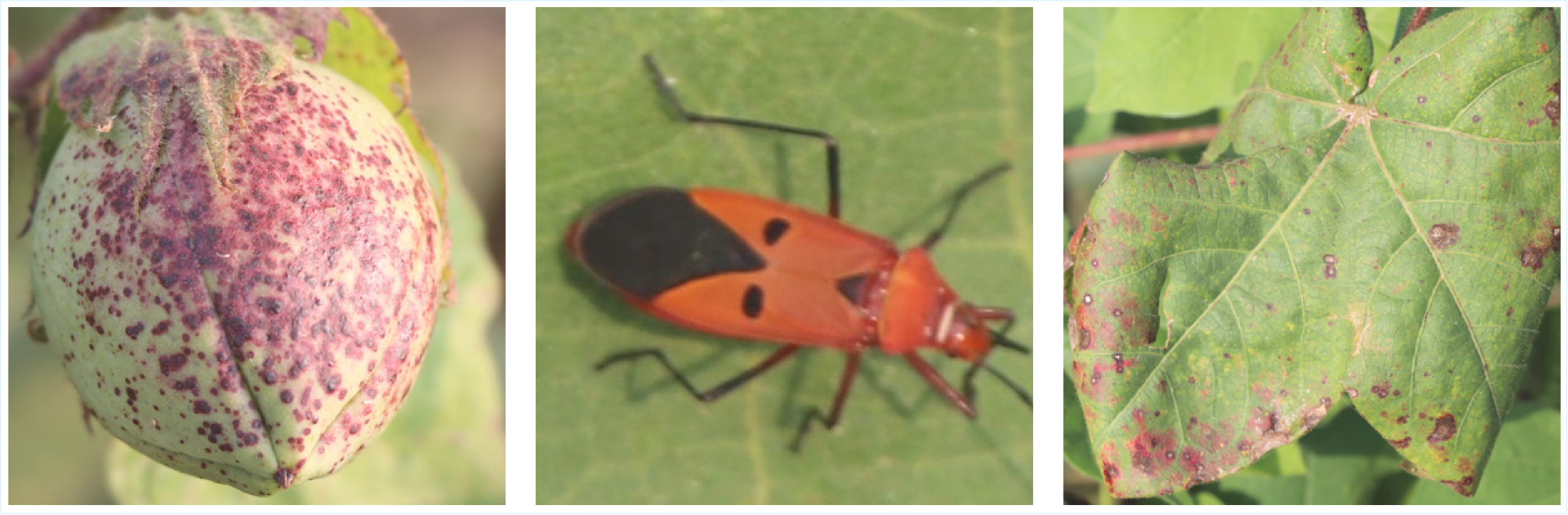}
  \caption{Disease classification results for cotton crops.The
image displays various symptoms, including a red cotton boll,
a red bug, and a leaf spot, representing different disease indicators identified by the classification model. }
    \label{fig:DSLR_images_sample_10}
\end{figure}

\noindent \textbf{Disease Classification.} The dataset is organized into three main categories: \textit{Leaf Diseases} (Fresh Leaf, Yellowish Leaf, Leaf Reddening, Leaf Spot/Bacterial Blight), \textit{Bugs} (Mealy Bug, Red Cotton Bug), and \textit{Cotton Boll} (Damaged Cotton Boll, Cotton Boll Rot, Healthy Cotton Boll). It was split into a 70:15:15 train:validation:test ratio for each class. The VGG19~\cite{mascarenhas2021comparison} model achieved 83.37\% test accuracy (Fig.~\ref{fig:DSLR_images_sample_10}) with training and validation losses of 0.3357 and 0.8805, respectively.

\vspace{-.3cm}

\subsection{Image Synthesis}
\vspace{-.2cm}
\begin{table}[!h]
    \centering
    \resizebox{0.49\textwidth}{!}{ % Slightly more compact than 1.0 but still readable
        \begin{tabular}{l c c | c c c}  % Vertical line separating the two sections
            \toprule
            \multicolumn{3}{c|}{\textbf{StyleGAN2-ADA(Dslr data)}} & \multicolumn{3}{c}{\textbf{StyleGAN2-ADA(Drone data)}} \\
            \midrule
            Metric & Value & Time (s) & Value & Time (s) \\
            \midrule
            IS (\(\mu \pm \sigma\)) & 3.214 ± 0.027 & 727.63 & 2.298 ± 0.009 & 726.25 \\
            FID & 36.44 & 761.79 & 30.9 & 761.99 \\
            KID & 0.0115 & 738.38 & 0.0077 & 734.95 \\
            \bottomrule
        \end{tabular}
    }
    \caption{The table compares performance metrics for StyleGAN2-ADA on cotton datasets captured by DSLR and Drone cameras. It reports Inception Score (IS), Fréchet Inception Distance (FID), and Kernel Inception Distance (KID), along with the associated computational time (in seconds).}
    \label{tab:combined_cotton_metrics}
\end{table}

\begin{figure}[!h]
  \includegraphics[width=\linewidth]{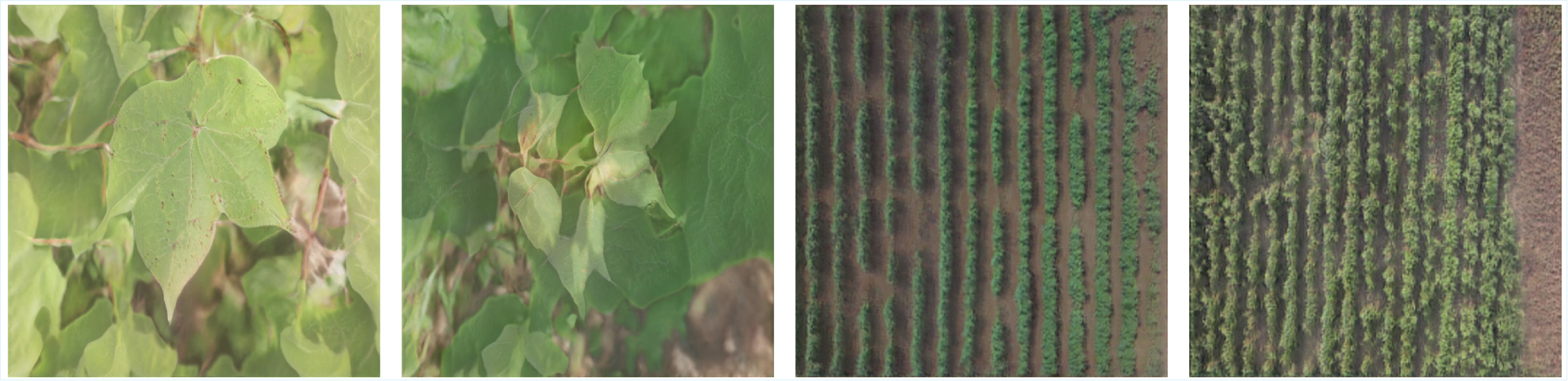}
      \caption{Synthetic images created using StyleGAN2-ADA: The first two images are derived from DSLR data, while the last two originate from drone data.}
    \label{fig:DSLR_images_sample_133}
\end{figure}

We used StyleGAN2-ADA for image synthesis, data augmentation, and conditional synthesis on drone and DSLR datasets, as given in Fig.~\ref{fig:DSLR_images_sample_133}. We also explored StyleGAN3 for improved image generation, benefiting from its advanced features. The matrices for drone and DSLR Data used in the synthesis process are presented in Table~\ref{tab:combined_cotton_metrics}, showing key parameters and metrics that assess the quality of the generated data.

% \begin{figure}[!h]
%     \centering
%     \begin{minipage}{0.24\linewidth}
%          \centering             
%          \includegraphics[width=0.99\linewidth]{images/synthesis/13.JPG}
%     \end{minipage}  
%     \begin{minipage}{0.24\linewidth}
%          \centering             
%          \includegraphics[width=0.99\linewidth]{images/synthesis/14.JPG}
%     \end{minipage}  
%     \begin{minipage}{0.24\linewidth}
%          \centering             
%          \includegraphics[width=0.99\linewidth]{images/synthesis/12.png}
%     \end{minipage}  
%     \begin{minipage}{0.24\linewidth}
%          \centering             
%          \includegraphics[width=0.99\linewidth]{images/synthesis/11.png}
%     \end{minipage}  
%     \caption{Synthetic images created using StyleGAN2-ADA: The first two images are derived from DSLR data, while the last two originate from drone data.}
%     \label{fig:DSLR_images_sample_133}
% \end{figure}

\begin{figure}
  \includegraphics[width=\linewidth]{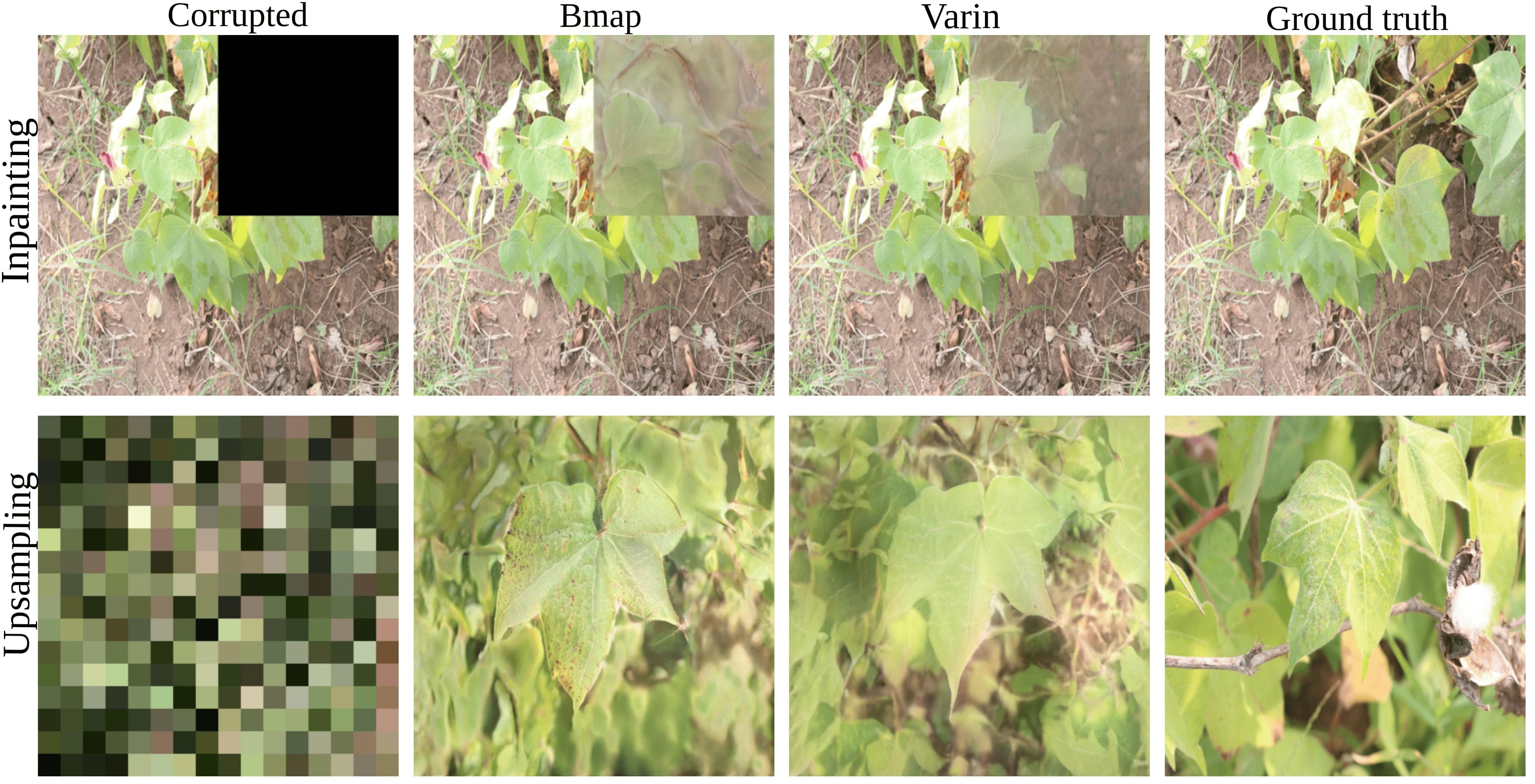}
      \caption{BGRM~\cite{marinescu2020bayesian} restoration using BMAP and Variational Inpainting (VarIn) for inpainting and super-resolution on the DSLR dataset.}
    \label{fig:DSLR_images_sample_13}
\end{figure}

\begin{table}[!ht]
    \centering
    \resizebox{0.49\textwidth}{!}{ 
        \begin{tabular}{p{35pt}p{28pt}p{28pt}p{28pt} p{28pt}p{28pt}p{28pt}} 
            \toprule
            Method & \multicolumn{3}{c|}{\textbf{Drone}} & \multicolumn{3}{c}{\textbf{DSLR}} \\
            \cline{2-7}
            & LPIPS↓ & LPIPS-vgg↓ & PFID↓ & LPIPS↓ & LPIPS-vgg↓ & PFID↓ \\
            \toprule
            \multicolumn{7}{l}{\textbf{Inpainting}} \\
            Bmap  & 0.214 & 0.198 & 59.67 & 0.214 & 0.212 & 70.78 \\
            Varin  & 0.203 & 0.194 & 51.57 & 0.210 & 0.208 & 62.07 \\
            \addlinespace[0.2em]
            \multicolumn{7}{l}{\textbf{Upsampling}} \\
            Bmap  & 0.678 & 0.643 & 135.13 & 0.656 & 0.669 & 135.32 \\
            Varin  & 0.674 & 0.649 & 143.03 & 0.655 & 0.661 & 128.46 \\
            \bottomrule
        \end{tabular}
    }
    \caption{Performance of Bayesian Reconstruction through Generative Models (BRGM)~\cite{marinescu2020bayesian} for inpainting and upsampling on Drone and DSLR cotton crop datasets.}
    \label{tab:BGRM}
\end{table}

\vspace{-.3cm}
\subsection{Image Restoration}
\vspace{-.1cm}
%We present the qualitative and quantitative results on restoring and enhancing the image quality of the proposed dataset using state-of-the-art deep-learning-based architectures.  
% \begin{figure}[!h]
%   \includegraphics[width=\linewidth]{images/restore.png}
%   \caption{Image Restoration Results .}
%   \label{fig:pred_comp_det}
% \end{figure}

% Image restoration is essential for addressing noise, artefacts, low resolution, and missing data in computer vision. We explored a robust unsupervised restoration~\cite{poirier2023robust}  framework using a StyleGAN-based architecture and the Bilateral Graph Regularization Model (BGRM) for improved restoration across degradation scenarios. \textcolor{red}{Evaluated on denoising, deartifacting, upsampling, and inpainting tasks with drone and DSLR image datasets, the results show significant quality improvements and state-of-the-art performance. The matrices for  BGMR are in Tables~\ref{tab:BGRM} for both.}

We restored the images of our proposed COT-AD dataset captured through a drone and DSLR using a robust unsupervised restoration~\cite{poirier2023robust} framework. To improve restoration across degradation scenarios, we performed deartifacting, upsampling, and inpainting tasks with drone and DSLR image datasets using a StyleGAN-based architecture and the Bilateral Graph Regularization Model (BGRM) (see Fig.~\ref{fig:DSLR_images_sample_13}). The matrices for  BGMR are represented in Table~\ref{tab:BGRM}.
\vspace{-.2cm}
\section{Conclusion}
\vspace{-.1cm}
In this paper, we have proposed a COT-AD dataset that significantly contributes to agricultural data resources, specifically tailored to cotton crop analysis. This dataset enables extensive applications such as classification, segmentation, and restoration by encompassing aerial and close-up images across various growth stages. Experimental evaluations have demonstrated that COT-AD supports accurate disease identification, efficient crop detection, and image restoration, proving valuable for both model training and real-world agricultural use cases. In future work, we propose the utilization of the dataset for the cotton-ball counting problem. 
\vspace{-.1cm}
\section{Acknowledgment}
\vspace{-.1cm}
We would like to acknowledge the support of the Jibaben Patel Chair
in AI,  L\&T Technology Services Limited, and Visvesvaraya PhD Scheme for completion of this work.

\vspace{-.3cm}

\bibliographystyle{IEEEbib}
\bibliography{strings}

\clearpage
\twocolumn[
\begin{center}
    \section*{Supplementary Material}
\end{center}
\vspace{1em}
]

\section{Dataset Documentation}

\begin{itemize}[label={}, leftmargin=30pt]
    \item \textbf{Description} \\
    This dataset comprises high-quality aerial and DSLR images designed for advanced agricultural applications, particularly targeting cotton crops. The aerial images offer extensive coverage suitable for detection and segmentation tasks, aiding in the analysis of crop growth and land use efficiency. The DSLR images are specifically curated for detailed classification of crop diseases, promoting early detection and management.
    \item \textbf{Dataset Link} \\\url{https://ieee-dataport.org/documents/cot-adcotton-analysis-dataset}
    \\\url{https://www.kaggle.com/datasets/aamaanakbar/cot-ad1}
    % \\\href{https://kaggle.com/datasets/8575ab4eee8f9286ef92ad2cf65eafdd2a4adf1649f93737c7b04e0619a3575a}{Click here to access the CottonDAD(annotated only) dataset.}
    \item \textbf{Data Subject(s)} \\ 
    Detection, Segmentation, Disease Detection, Restoration, Synthesis
    \item \textbf{Data Type}\\ Dataset consists of Aerial and DSLR Images along with videos of Cotton crops in two farms for the entire life cycle.
    \item \textbf{Volume}\\ Approximately 310 GB.
    \item \textbf{Total Number of Images}\\ The dataset comprises more than 25,000 images.
    \item \textbf{Total Number of Videos}\\ The dataset comprises more than 140  videos.
    \item \textbf{Image Format}\\ The images are in JPG,JPEG and mp4 formats.
    \item \textbf{Time Frame of Data Collection}\\ Collected over a full harvest cycle of cotton crops across two farms.
    \item \textbf{Data Source}\\ Aerial images captured through drones with an altitude of [10,15,115] Meters.\\ DSLR images taken in zoomed in mode.
    \item \textbf{Label Details for Aerial Images} \\Separate folders containing the images; \textbf{segmentation masks}, that is, the binary masks for crop segmentation in \textit{JPG} format; \textbf{segmentation labels} for crop segmentation in YOLO standard \textit{.txt} format; and \textbf{detection labels} for crop detection in YOLO standard \textit{.txt} format.
    \item \textbf{Label Details for DSLR Images}\\ Each disease type is organized into distinct folders containing the corresponding \textit{JPG} images.
    \item \textbf{Annotation Details} \\ 
    Each image underwent meticulous manual annotation followed by several rigorous rounds of quality checks to ensure high accuracy and reliability.
    \item \textbf{Dataset Owner and Affiliation} \\
    \href{https://www.iitgn.ac.in}{Indian Institute of Technology Gandhinagar,  Gujarat, INDIA}
    \item \textbf{Version Details} \\ Current Version is 1.0
    \item \textbf{Maintenance Status} \\
    Limited Maintenance - The data will not be updated, but any technical issues will be addressed.
    \item \textbf{Sample of Data} \\ Provided in \textbf{Figure 1} and \textbf{Figure 2} of main paper.
    \item \textbf{Dataset Organization} \\ The dataset organization is provided in Section~\ref{sec:data_org}
    % \item \textbf{Training and Validation Codes} \\ \href{https://iitgnacin-my.sharepoint.com/:f:/g/personal/22310050_iitgn_ac_in/EtdFSWtLYFVGtqN0ixOGIawBFnbe1D5n5pyBDbQJXBGPdQ?e=LHah2H}{Click here to access the training and validation codes}.
    \item \textbf{License} \\ ATTRIBUTION-NONCOMMERCIAL 4.0 INTERNATIONAL
    \item \textbf{DOI} \\ 10.21227/bpqn-9a12(ieee dataport) AND 10.34740/kaggle/dsv/10674370(kaggle)
    % \item \textbf{Meta-Data} \\ \href{https://drive.google.com/drive/folders/1AEFxUDVQ7TeY4Ef3yFhLYNK137a0m80l?usp=sharing}{Click here to to access the meta-data}.
    \item \textbf{Author statement} \\ We state that we bear all responsibility in case of violation of rights, etc.
    
\end{itemize}

\section{Need of cotton data}
Agriculture and food management are critical challenges that the world will face in the near future. Advancements in crop management and sustainable practices will be essential to ensure adequate food supplies. Utilizing aerial vehicles and cameras combined with advanced computer vision techniques can address various aspects of crop management, such as disease detection, weed density analysis, and harvesting information, enabling farmers to make timely and informed decisions. This paper presents two comprehensive datasets focused on cotton crops: an aerial dataset capturing the crop's entire life cycle and a dataset of up-close DSLR images detailing various diseases affecting cotton crops at different stages. The aerial dataset is meticulously annotated for cotton crop detection and segmentation, offering valuable insights into vegetation and weed competition in specific farm areas. The DSLR dataset includes images of leaf diseases, cotton boll rot, and red cotton bugs, facilitating the classification and early management of cotton diseases. 

Our dataset serves a key gap in the existing agricultural databases by offering fine-grained segmentation data specifically optimized for cotton crops, making it more useful for plant-level as well as field-level detection and segmentation tasks. With over 5000 images that have been hand-annotated, taken weekly throughout an entire harvest cycle, it is among the most extensive datasets present. These data sets can be used as baselines to create innovative technological solutions to enhance agricultural practices, opening the door to more efficient and sustainable agriculture.Food management and agriculture are vital issues the world will encounter in the near future.

% \usepackage{graphicx}
\begin{table*}[!t]
\centering
\resizebox{\textwidth}{!}{%
\begin{tabular}{lll}
\hline
\multicolumn{1}{c}{\textbf{Month \#}} &
  \multicolumn{1}{c}{\textbf{Stage of Cotton}} &
  \multicolumn{1}{c}{\textbf{Interpretation}} \\ \hline
Month 0 &
  Root establishment &
  Plantation and cotton plants started growing. \\ \hline
Month 1 &
  Leaf Area Expansion and Flowering &
  \begin{tabular}[c]{@{}l@{}}Plants were growing and leaf area increased. In the \\ last week, flowering started in some plants.\end{tabular} \\ \hline
Month 2 &
  Flowering + cotton boll development &
  \begin{tabular}[c]{@{}l@{}}Flower development in good amount. Cotton boll \\ development. Very few cotton bolls started opening.\end{tabular} \\ \hline
Month 3 &
  \begin{tabular}[c]{@{}l@{}}Flowering + cotton boll development \\ + cotton boll opening\end{tabular} &
  Flowering and cotton boll development increased. \\ \hline
Month 4 &
  \begin{tabular}[c]{@{}l@{}}Flowering + cotton boll development \\ + cotton boll opening and harvesting\end{tabular} &
  \begin{tabular}[c]{@{}l@{}}Flowering and cotton boll development seen across\\ the farm. In some farm areas, plants were seen \\ moving towards the end of the life cycle.\end{tabular} \\ \hline
Month 5 &
  \begin{tabular}[c]{@{}l@{}}Flowering + cotton boll development \\ + cotton boll opening and harvesting\end{tabular} &
  \begin{tabular}[c]{@{}l@{}}Cotton bolls are harvested in between. Many plants \\ move toward the end of their life cycles.\end{tabular} \\ \hline
Month 6 &
  Cotton Boll opening and harvesting &
  \begin{tabular}[c]{@{}l@{}}New cotton flowering stops. The rest of the bolls are \\ developing and opening, and many plants are in \\ the final stages of their life cycle.\end{tabular} \\ \hline
\end{tabular}%
}
\caption{Month-wise interpretation of cotton farms for aerial imagery}
\label{table:month_wise}
\end{table*}

% \usepackage{graphicx}
\begin{table*}[!t]
\centering
\resizebox{\textwidth}{!}{%
\begin{tabular}{llll}
\hline
\textbf{Month} &
  \textbf{Stage of Cotton} &
  \textbf{Diseases Seen} &
  \textbf{Interpretation} \\ \hline
Month 2 &
  \begin{tabular}[c]{@{}l@{}}Flowering + cotton boll \\ development\end{tabular} &
  Mealy Bug development &
  \begin{tabular}[c]{@{}l@{}}The plants are growing. Mealy bug infestation started \\ on some leaf and branches. Towards the end of the \\ month, yellowish-ness was visible in plants.\end{tabular} \\ \hline
Month 3 &
  \begin{tabular}[c]{@{}l@{}}Flowering + cotton boll \\ development + cotton \\ boll opening\end{tabular} &
  \begin{tabular}[c]{@{}l@{}}Mealy Bug + Leaf Disease \\ symptoms + some cotton bolls \\ started having pink spots\end{tabular} &
  \begin{tabular}[c]{@{}l@{}}Mealy bug infestation found in both the farms. \\ The leaf diseases started growing. Spots started \\ developing in some leaves. Some cotton bolls \\ started having pinkish spots.\end{tabular} \\ \hline
Month 4 &
  \begin{tabular}[c]{@{}l@{}}Flowering + cotton boll \\ development + cotton boll \\ opening and harvesting\end{tabular} &
  \begin{tabular}[c]{@{}l@{}}Leaf Disease + Pink Cotton \\ Boll Rot + Red Cotton Bug\end{tabular} &
  \begin{tabular}[c]{@{}l@{}}Leaf disease and infection increased with time. \\ Cotton Boll pink spot increased. Mealy Bug numbers \\ dropped. Red Cotton Bugs presence developed.\end{tabular} \\ \hline
Month 5 &
  \begin{tabular}[c]{@{}l@{}}Flowering + cotton boll \\ development + cotton boll \\ opening and harvesting\end{tabular} &
  \begin{tabular}[c]{@{}l@{}}Leaf Disease + Pink Cotton \\ Boll Rot + Red Cotton Bug \\ + Leaf Reddening\end{tabular} &
  \begin{tabular}[c]{@{}l@{}}Same as above. Additionally, Leaf reddening \\ symptoms started to increase. Red cotton Bugs\\  infestation decreased with time.\end{tabular} \\ \hline
Month 6 &
  \begin{tabular}[c]{@{}l@{}}Cotton Boll opening and\\ harvesting\end{tabular} &
  \begin{tabular}[c]{@{}l@{}}Leaf Disease + Pink Cotton \\ Boll Rot + Leaf Reddening\end{tabular} &
  \begin{tabular}[c]{@{}l@{}}Leaf disease, leaf reddening and cotton boll rotting\\ increased, many cotton bolls were dropped before \\ maturity. Till the end, all the plants were infected.\end{tabular} \\ \hline
\end{tabular}%
}
\caption{Month-wise interpretation of cotton farm for disease classification}
\label{table:month_wise_disease}
\end{table*}

% Our dataset fills a crucial gap in the current agricultural repositories by providing detailed segmentation data tailored for cotton crops, enhancing its utility for both plant-level and field-level detection and segmentation tasks. With more than 5000 manually annotated images, captured weekly over a complete harvest cycle, it is one of the most comprehensive datasets available. These datasets can serve as benchmarks for developing advanced technical solutions to improve agricultural practices, paving the way for more efficient and sustainable farming.
% Agriculture and food management are critical challenges that the world will face in the near future. 
Advancements in crop management and sustainable practices will be essential to ensure adequate food supplies. Utilizing aerial vehicles and cameras combined with advanced computer vision techniques can address various aspects of crop management, such as disease detection, weed density analysis, and harvesting information, enabling farmers to make timely and informed decisions regarding crop management. This paper presents an aerial dataset of cotton fields captured throughout the crop's life cycle and a dataset with up-close DSLR images of various diseases affecting cotton crops at different life cycle stages.

\begin{table*}[!t]
    \centering
    \resizebox{0.75\textwidth}{!}{
            \begin{tabular}{p{40pt}p{30pt}p{30pt}p{30pt}|p{30pt}p{30pt}p{30pt}} % Vertical line added here
            \toprule
            Method & \multicolumn{3}{c|}{\textbf{DSLR Cotton Crop Data}} & \multicolumn{3}{c}{\textbf{Drone Cotton Crop Data}} \\
            \cline{2-7}
            & LPIPS↓ & LPIPS-vgg↓ & PFID↓ & LPIPS↓ & LPIPS-vgg↓ & PFID↓ \\
            \toprule
            \multicolumn{7}{c}{\textbf{Deartifacting (JPEG Compression)}} \\
            L & 0.584 & 0.619 & 121.419 & 0.598 & 0.588 & 120.462 \\
            M & 0.576 & 0.613 & 115.922 & 0.583 & 0.580 & 112.291 \\
            S & 0.573 & 0.611 & 114.228 & 0.579 & 0.576 & 113.928 \\
            XL & 0.6   & 0.633 & 118.723 & 0.615 & 0.601 & 157.034 \\
            XS & 0.571 & 0.609 & 115.213 & 0.577 & 0.575 & 113.201 \\
            \addlinespace[0.2em]
            \multicolumn{7}{c}{\textbf{Denoising (Clamped Poisson and Bernoulli Mixture)}} \\
            L & 0.58  & 0.628 & 116.625  & 0.576 & 0.583 & 130.393 \\
            M & 0.573 & 0.62  & 113.264 & 0.570 & 0.579 & 115.636 \\
            S & 0.567 & 0.613 & 109.726 & 0.568 & 0.575 & 114.738 \\
            XL & 0.592 & 0.64  & 115.037 & 0.591 & 0.591 & 146.684 \\
            XS & 0.56  & 0.607 & 113.883 & 0.562 & 0.571 & 107.531 \\
            \addlinespace[0.2em]
            \multicolumn{7}{c}{\textbf{Inpainting (Random Strokes)}} \\
            L & 0.553 & 0.596 & 126.934 & 0.527 & 0.553 & 112.856 \\
            M & 0.548 & 0.59  & 121.841 & 0.528 & 0.552 & 106.999 \\
            S & 0.542 & 0.585 & 114.523 & 0.524 & 0.549 & 104.970 \\
            XL & 0.558 & 0.602 & 133.558 & 0.533 & 0.557 & 109.472 \\
            XS & 0.551 & 0.59  & 114.447 & 0.543 & 0.556 & 96.843  \\
            \addlinespace[0.2em]
            \multicolumn{7}{c}{\textbf{Upsampling (Bilinear, Bicubic or Lanczos)}} \\
            L & 0.619 & 0.642 & 129.074 & 0.604 & 0.599 & 154.395 \\
            M & 0.598 & 0.632 & 126.482 & 0.606 & 0.598 & 140.140 \\
            S & 0.581 & 0.619 & 119.499 & 0.592 & 0.587 & 103.984 \\
            XL & 0.628 & 0.651 & 130    & 0.620 & 0.609 & 161.933 \\
            XS & 0.566 & 0.605 & 116.894 & 0.578 & 0.577 & 102.341 \\
            \addlinespace[0.2em]
            \multicolumn{7}{c}{\textbf{2 Degradations}} \\
            AP & 0.57  & 0.61  & 121.229 & 0.542 & 0.569 & 112.517 \\
            NA & 0.584 & 0.627 & 114.7   & 0.584 & 0.589 & 136.752 \\
            NP & 0.57  & 0.615 & 122.015 & 0.538 & 0.566 & 118.027 \\
            UA & 0.639 & 0.673 & 127.588 & 0.636 & 0.617 & 163.628 \\
            UN & 0.623 & 0.66  & 116.295 & 0.604 & 0.605 & 140.503 \\
            UP & 0.588 & 0.626 & 120.72  & 0.573 & 0.583 & 139.341 \\
            \addlinespace[0.2em]
            \multicolumn{7}{c}{\textbf{3 Degradations}} \\
            NAP & 0.578 & 0.621 & 122.239 & 0.542 & 0.569 & 123.345 \\
            UAP & 0.617 & 0.66  & 124.197 & 0.592 & 0.596 & 119.654 \\
            UNA & 0.644 & 0.68  & 119.279 & 0.614 & 0.612 & 138.181 \\
            UNP & 0.624 & 0.654 & 124.385 & 0.557 & 0.578 & 108.822 \\
            \addlinespace[0.2em]
            \multicolumn{7}{c}{\textbf{4 Degradations}} \\
            UNAP & 0.638 & 0.67  & 120.707 & 0.598 & 0.591 & 107.681 \\
            \bottomrule
        
        \end{tabular}
    }
\caption{Performance matrics of the RUSIR\cite{poirier2023robust}  for image restoration on cotton crop data from DSLR and Drone cameras, evaluated across five levels (S, M, L, XL, XS) for single degradation types (deartifacting, denoising, inpainting, and upsampling), we also used multiple degradations. The RUSIR method uses a consistent set of hyperparameters across all levels, while baselines are optimized for accuracy at each individual level. The reported metrics include Accuracy (LPIPS), Fidelity (LPIPS-vgg), and Realism (PFID), with lower values indicating better performance.}
\label{table:RUSIR_mul}
\end{table*}

\section{Methodology }
\textbf{Harnessing Crops Detection in Modern Farming.} In smart farming, accurately detecting crops is essential for enhancing agricultural productivity and sustainability. Our dataset facilitates precise crop detection and is foundational for numerous smart farming applications. 
The technology enables crop health and growth to be monitored using drones with object detection features. Detection at an early stage helps to apply particular interventions, which in turn help in attaining healthier crop harvests. Accurate identification of crops is important to optimizing the use of farm inputs such as water, fertilizers, and pesticides. With precise specification of the location and condition of crops, farmers are able to use these inputs more effectively and with reduced wastage and degradation of the environment. Not only does precision application save resources, but also the health of crops is enhanced by the application of interventions specifically where and when needed. Crop identification is extremely important in making the processes of automating and streamlining cotton bud harvesting processes possible. With accurate identification of mature crops and buds, automatic systems can efficiently harvest them at their peak ripeness, reducing labour costs and crop as well as soil damage.

\begin{figure}[!h]
  \includegraphics[width=\linewidth]{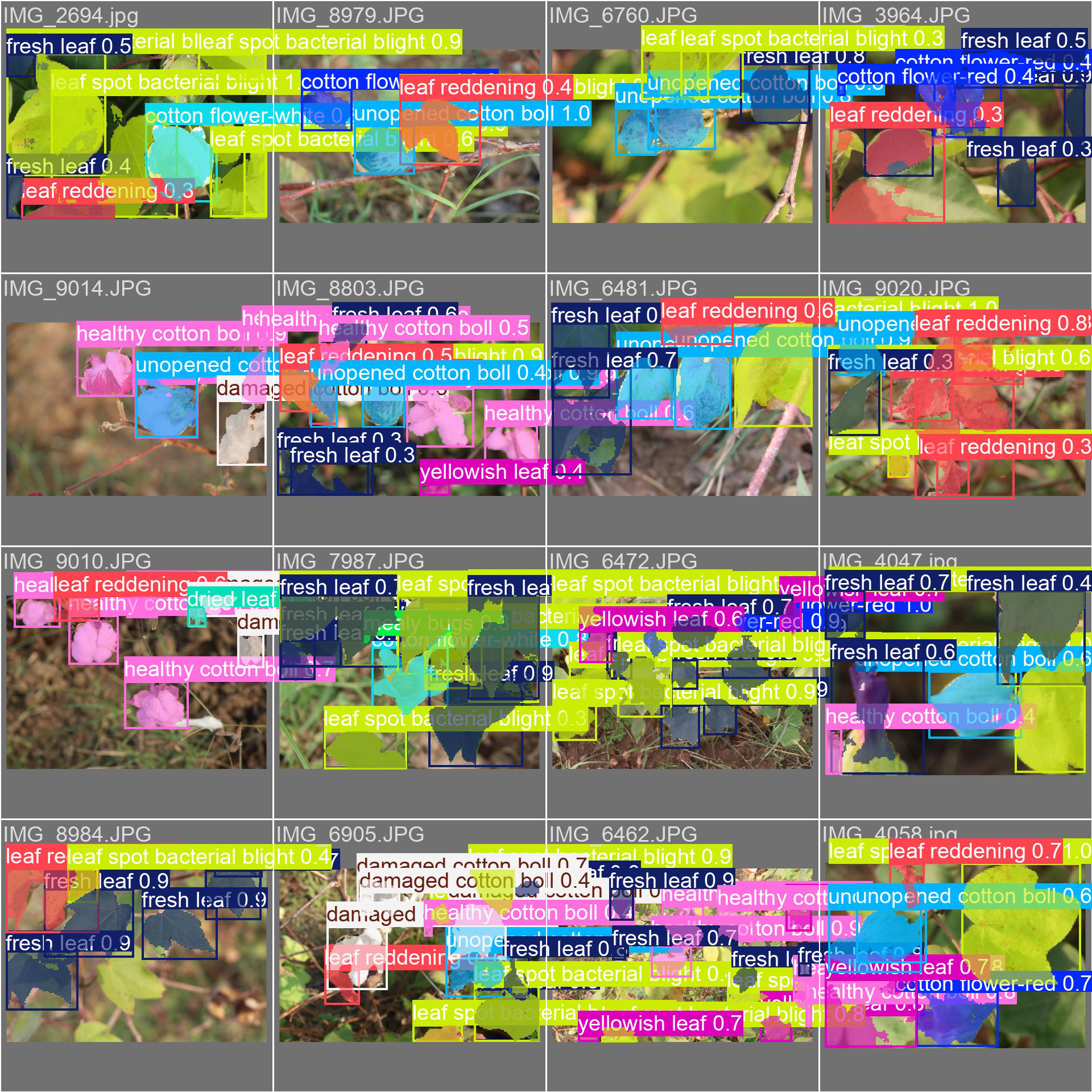}
  \caption{Cotton Class identification result using yolov11.  }
  \label{fig:seg_performancebghg}
\end{figure}

\begin{figure*}[!h]
    \centering

    % ---------- First row: labels ----------
    \begin{minipage}{0.24\linewidth}
        \centering
        \scriptsize (a) Input Image
    \end{minipage}
    \begin{minipage}{0.24\linewidth}
        \centering
        \scriptsize (b) LDAST~\cite{ldast_ECCV2022}
    \end{minipage}
    \begin{minipage}{0.24\linewidth}
        \centering
        \scriptsize (c) MMIST~\cite{wang2024multimodality}
    \end{minipage}
    \begin{minipage}{0.24\linewidth}
        \centering
        \scriptsize (d) CLIPtone~\cite{Lee_2024_CVPR}
    \end{minipage}

    % ---------- Second row ----------
    \begin{minipage}{0.24\linewidth}
        \centering
        \includegraphics[width=\linewidth]{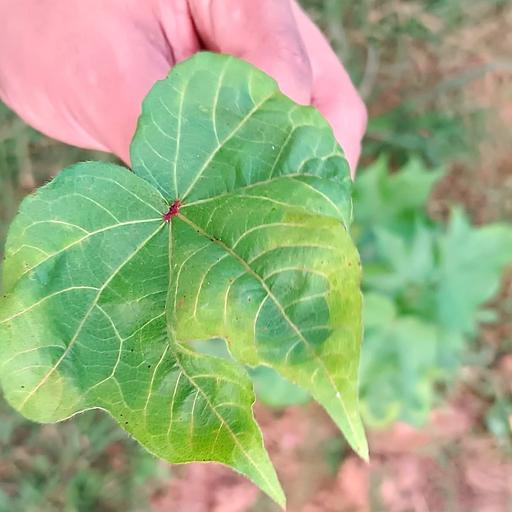}
    \end{minipage}
    \begin{minipage}{0.24\linewidth}
        \centering
        \includegraphics[width=\linewidth]{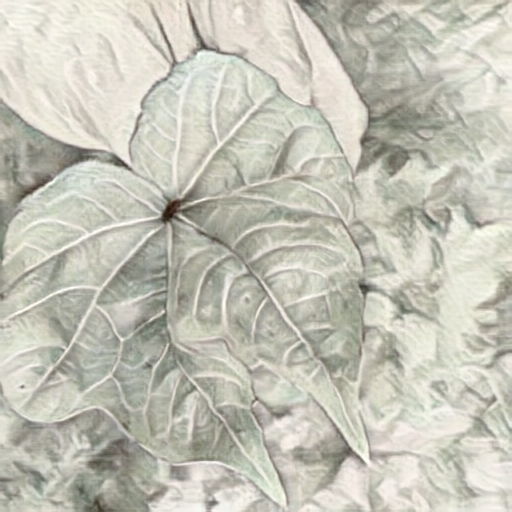}
    \end{minipage}
    \begin{minipage}{0.24\linewidth}
        \centering
        \includegraphics[width=\linewidth]{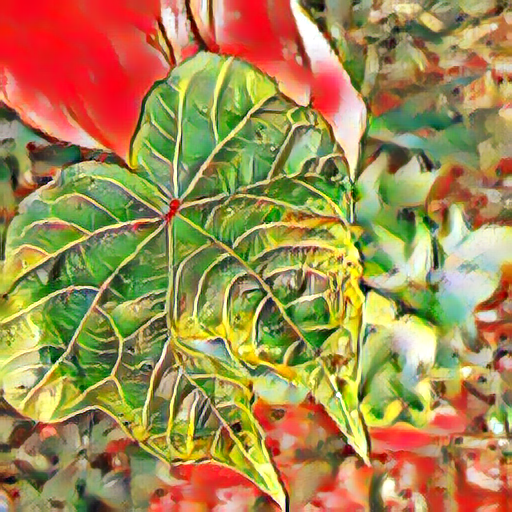}
    \end{minipage}
    \begin{minipage}{0.24\linewidth}
        \centering
        \includegraphics[width=\linewidth]{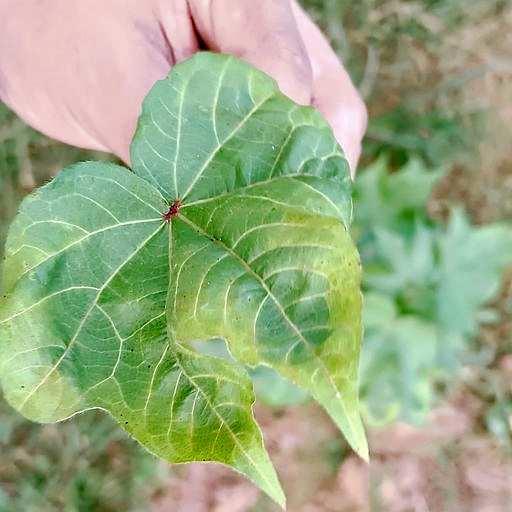}
    \end{minipage}

    % ---------- Third row ----------
    \begin{minipage}{0.24\linewidth}
        \centering
        \includegraphics[width=\linewidth]{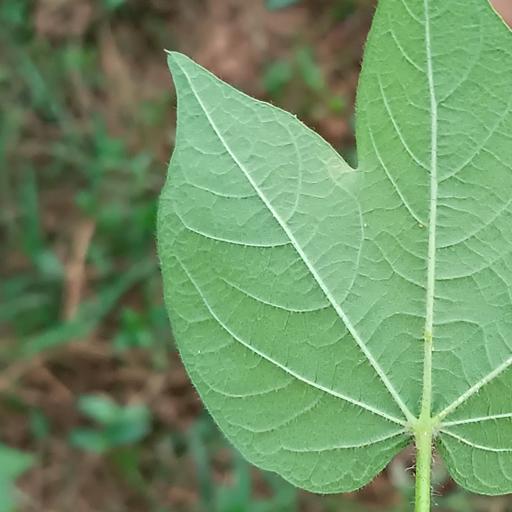}
    \end{minipage}
    \begin{minipage}{0.24\linewidth}
        \centering
        \includegraphics[width=\linewidth]{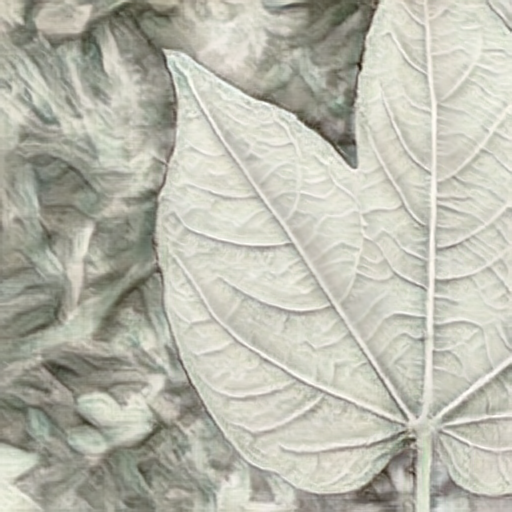}
    \end{minipage}
    \begin{minipage}{0.24\linewidth}
        \centering
        \includegraphics[width=\linewidth]{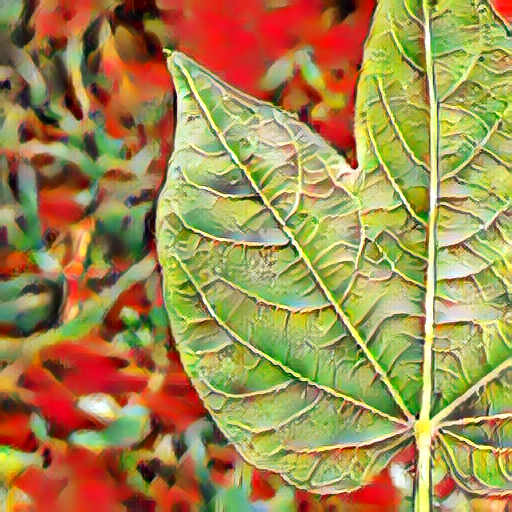}
    \end{minipage}
    \begin{minipage}{0.24\linewidth}
        \centering
        \includegraphics[width=\linewidth]{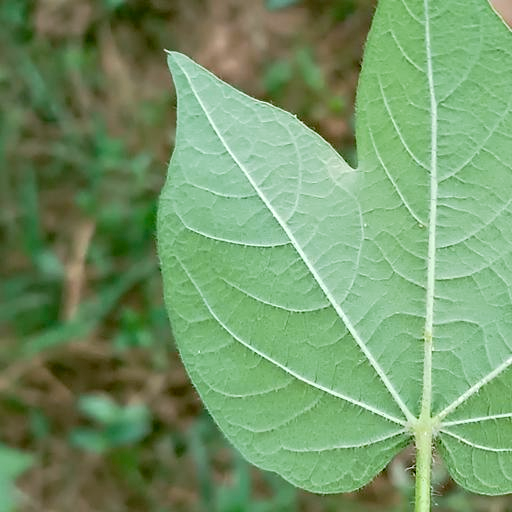}
    \end{minipage}

    % ---------- Fourth row ----------
    \begin{minipage}{0.24\linewidth}
        \centering
        \includegraphics[width=\linewidth]{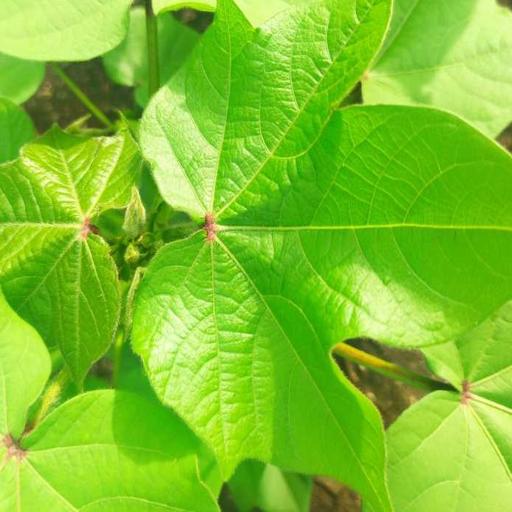}
    \end{minipage}
    \begin{minipage}{0.24\linewidth}
        \centering
        \includegraphics[width=\linewidth]{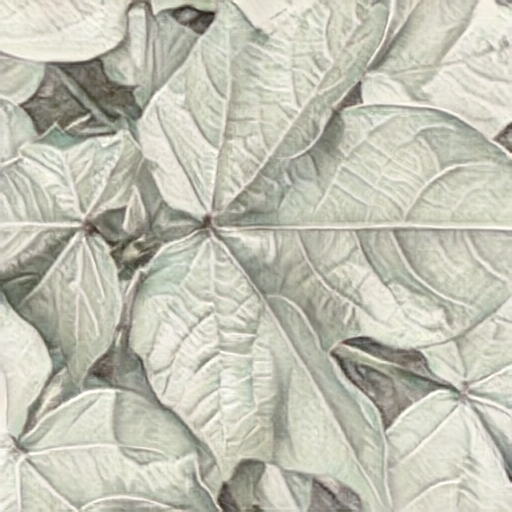}
    \end{minipage}
    \begin{minipage}{0.24\linewidth}
        \centering
        \includegraphics[width=\linewidth]{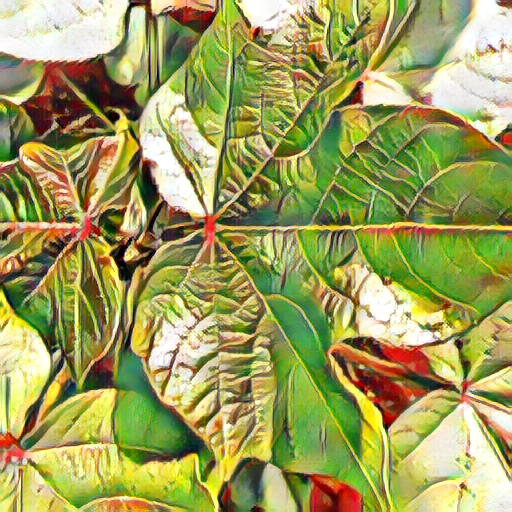}
    \end{minipage}
    \begin{minipage}{0.24\linewidth}
        \centering
        \includegraphics[width=\linewidth]{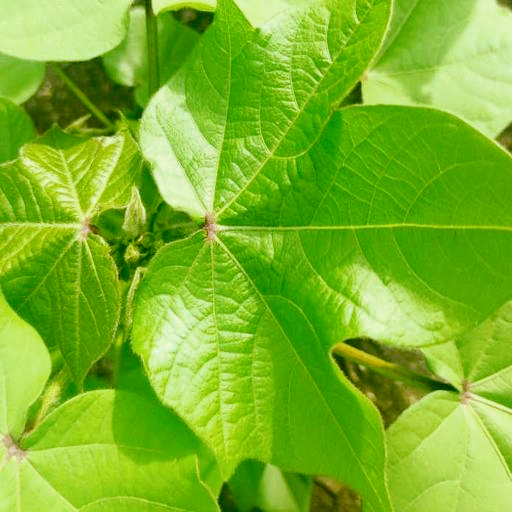}
    \end{minipage}

    % ---------- Fifth row ----------
    \begin{minipage}{0.24\linewidth}
        \centering
        \includegraphics[width=\linewidth]{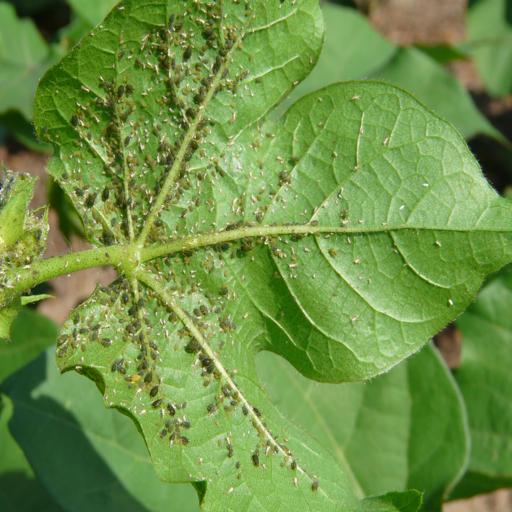}
    \end{minipage}
    \begin{minipage}{0.24\linewidth}
        \centering
        \includegraphics[width=\linewidth]{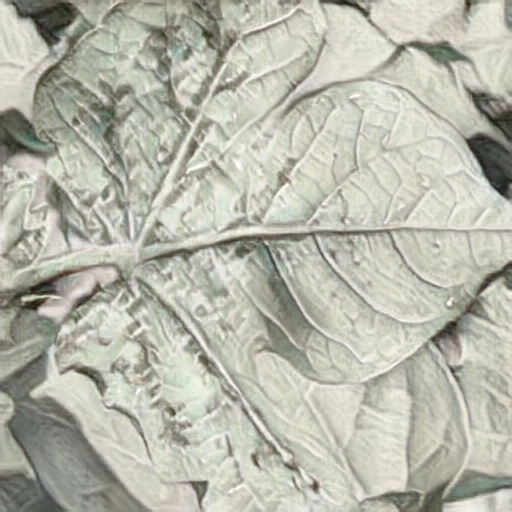}
    \end{minipage}
    \begin{minipage}{0.24\linewidth}
        \centering
        \includegraphics[width=\linewidth]{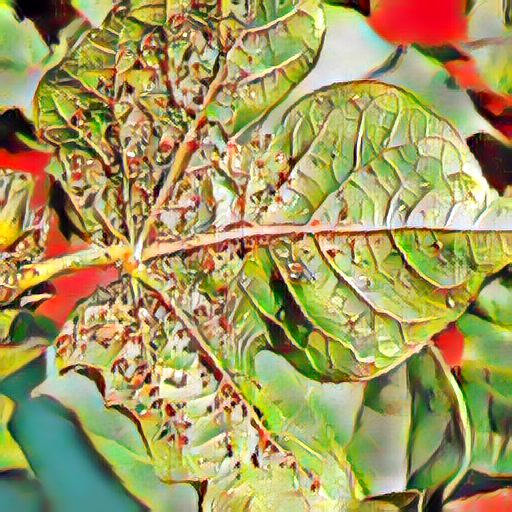}
    \end{minipage}
    \begin{minipage}{0.24\linewidth}
        \centering
        \includegraphics[width=\linewidth]{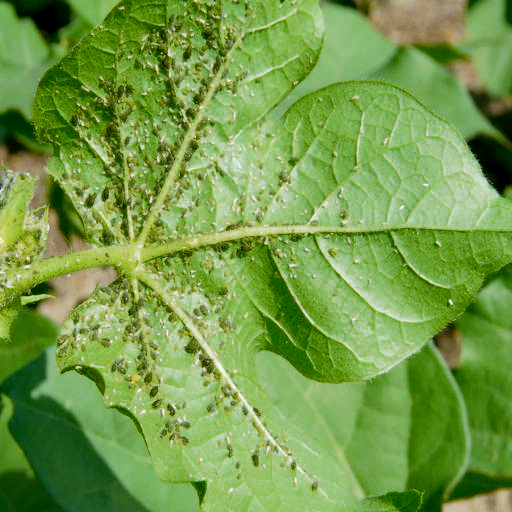}
    \end{minipage}

    \caption{\textbf{Image Enhancement Results.} Each row shows a different dataset(refer main paper table 1)(top to bottom: Augment(Cotton Leaf Disease Detection\cite{bishshash2024comprehensive}), Original(Cotton Leaf Disease Detection) Dataset\cite{bishshash2024comprehensive}, cotton leaf disease dataset\cite{cotton_leaf_disease_dataset} and A Dataset of Cotton Leaf Images for Disease Detection and Classification\cite{miranidataset}) images and enhancement results across four methods: LDAST\cite{ldast_ECCV2022}, MMIST\cite{wang2024multimodality} and CLiptone\cite{Lee_2024_CVPR}. \textbf{Prompt :} A clean, high-quality image with balanced contrast and brightness, free from noise or any distortion.}
    \label{fig:img_enhan_datasets}
\end{figure*}

% The primary benefit of crop segmentation is the accurate identification and delineation of individual plants within larger fields. This precision is crucial for effective crop yield estimation. By understanding the specific characteristics of each plant, such as size, density, and overall health, farmers can make accurate predictions about yield. This information helps in planning resource allocation and scheduling harvest operations more efficiently. Another critical application of crop segmentation is in the management of weeds. The ability to differentiate between crops and weeds within imagery enables targeted interventions. Farmers can apply herbicides or perform weed removal precisely where needed, significantly reducing the amount of chemicals used and minimizing their impact on the crops and environment. Disease and pest management also benefits greatly from accurate crop segmentation. Early detection of affected areas allows for targeted treatment, preventing disease spread and minimising crop yield damage. By isolating and treating only the affected areas, farmers can use pesticides more sparingly and effectively.

\begin{figure}[!h]
    \centering
    \begin{minipage}{0.24\linewidth}
    \centering
    \scriptsize
        Input \\ Image
    \end{minipage}
   \begin{minipage}{0.24\linewidth}
        \centering
     \scriptsize
         GT
     \end{minipage}
  \begin{minipage}{0.24\linewidth}
    \centering
    \scriptsize
        DSM~\cite{melas2022deep}
    \end{minipage}
  \begin{minipage}{0.24\linewidth}
    \centering
    \scriptsize
        SAM
    \end{minipage}
   \begin{minipage}{0.24\linewidth}
         \centering
             \includegraphics[width=0.99\linewidth]{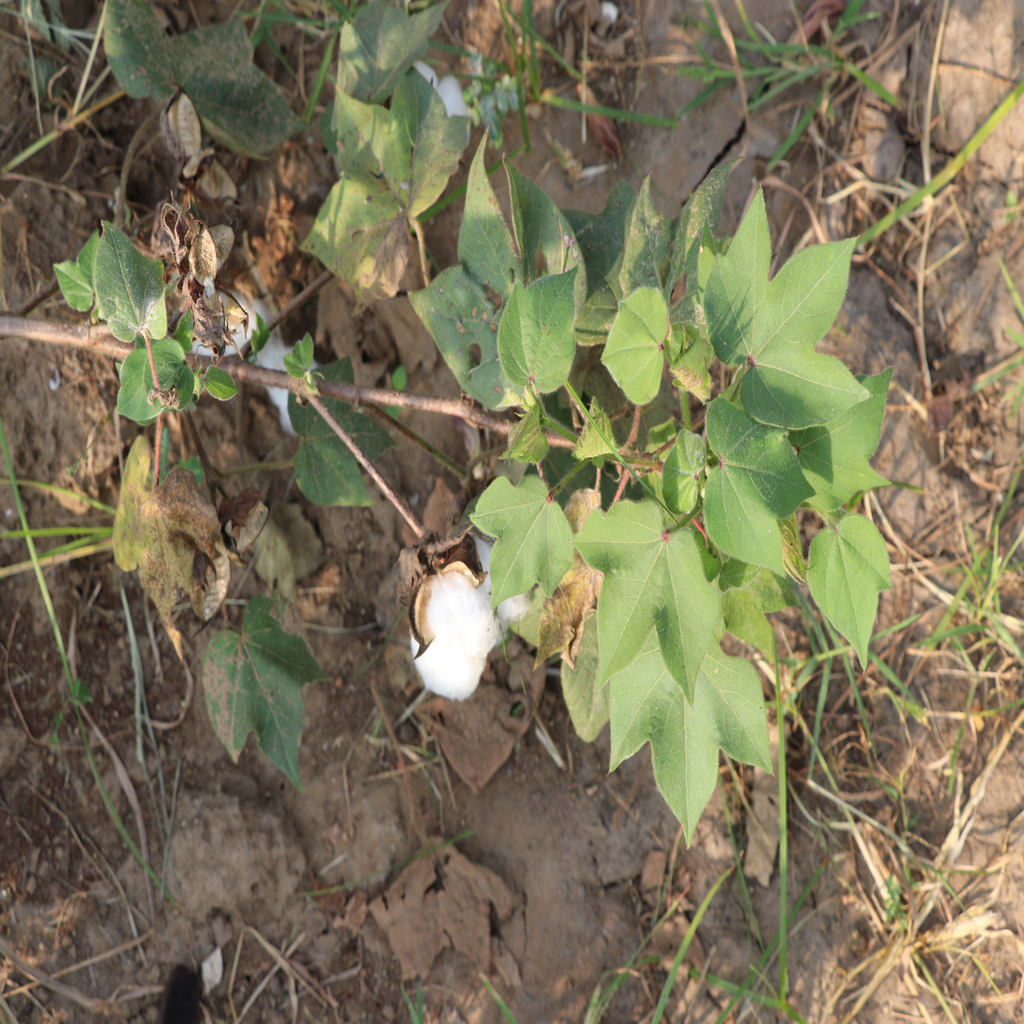}
        \end{minipage}  
   \begin{minipage}{0.24\linewidth}
         \centering
             \includegraphics[width=0.99\linewidth]{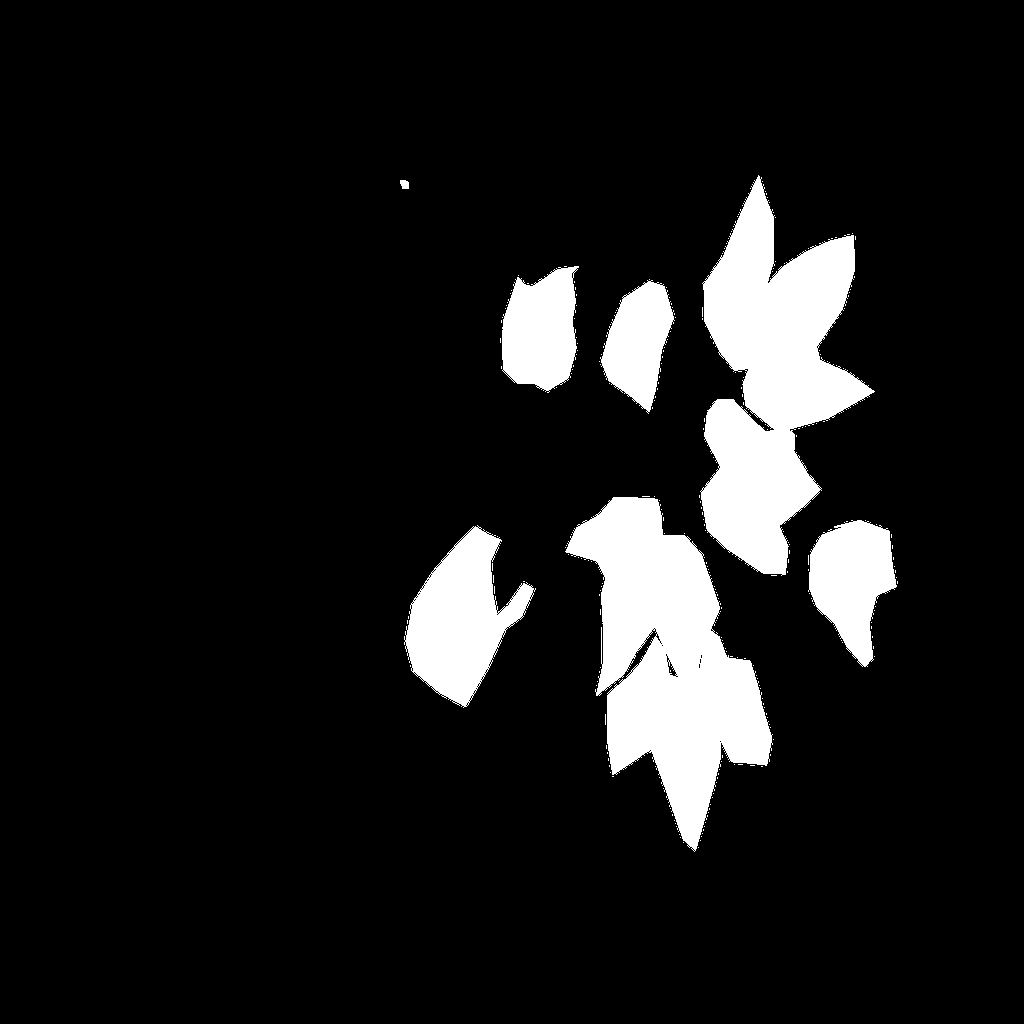}
    \end{minipage}  
   \begin{minipage}{0.24\linewidth}
         \centering
             \includegraphics[width=0.99\linewidth]{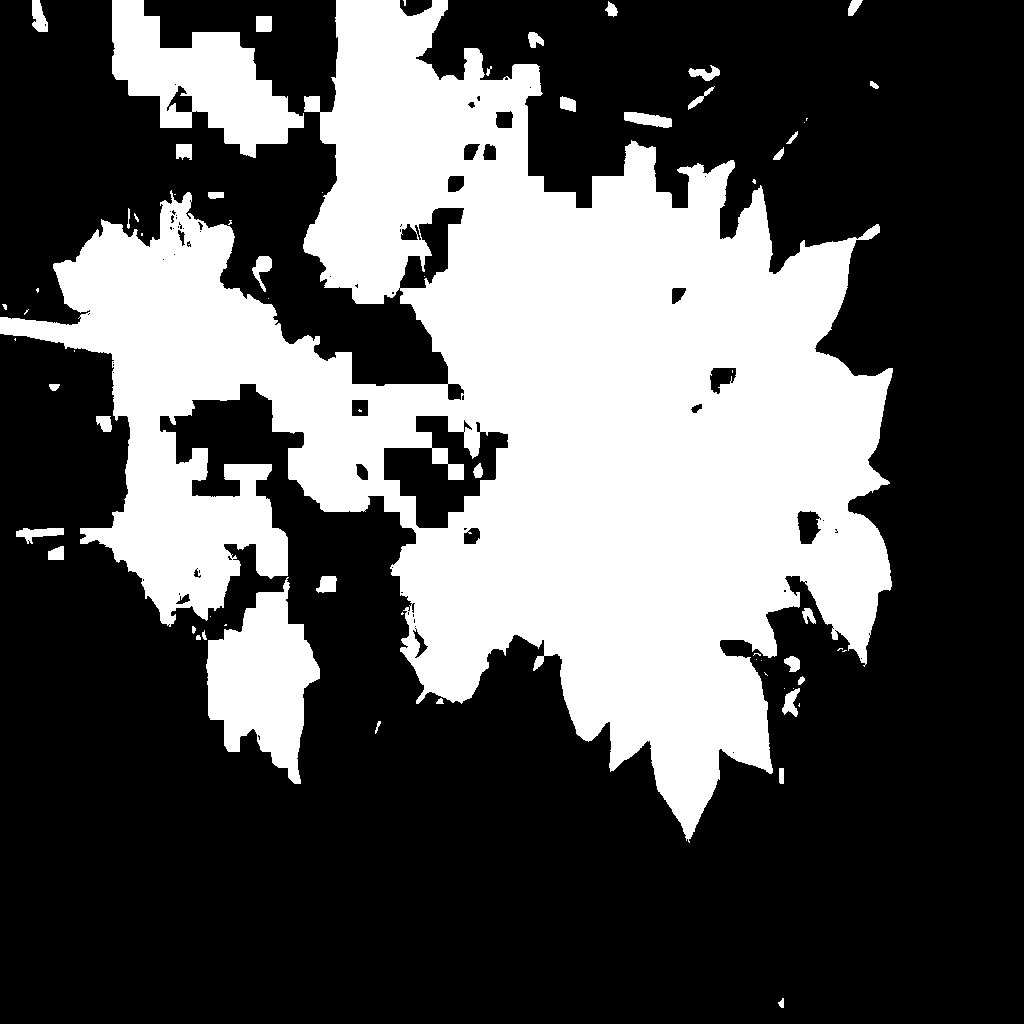}
    \end{minipage}  
      \begin{minipage}{0.24\linewidth}
         \centering
             \includegraphics[width=0.99\linewidth]{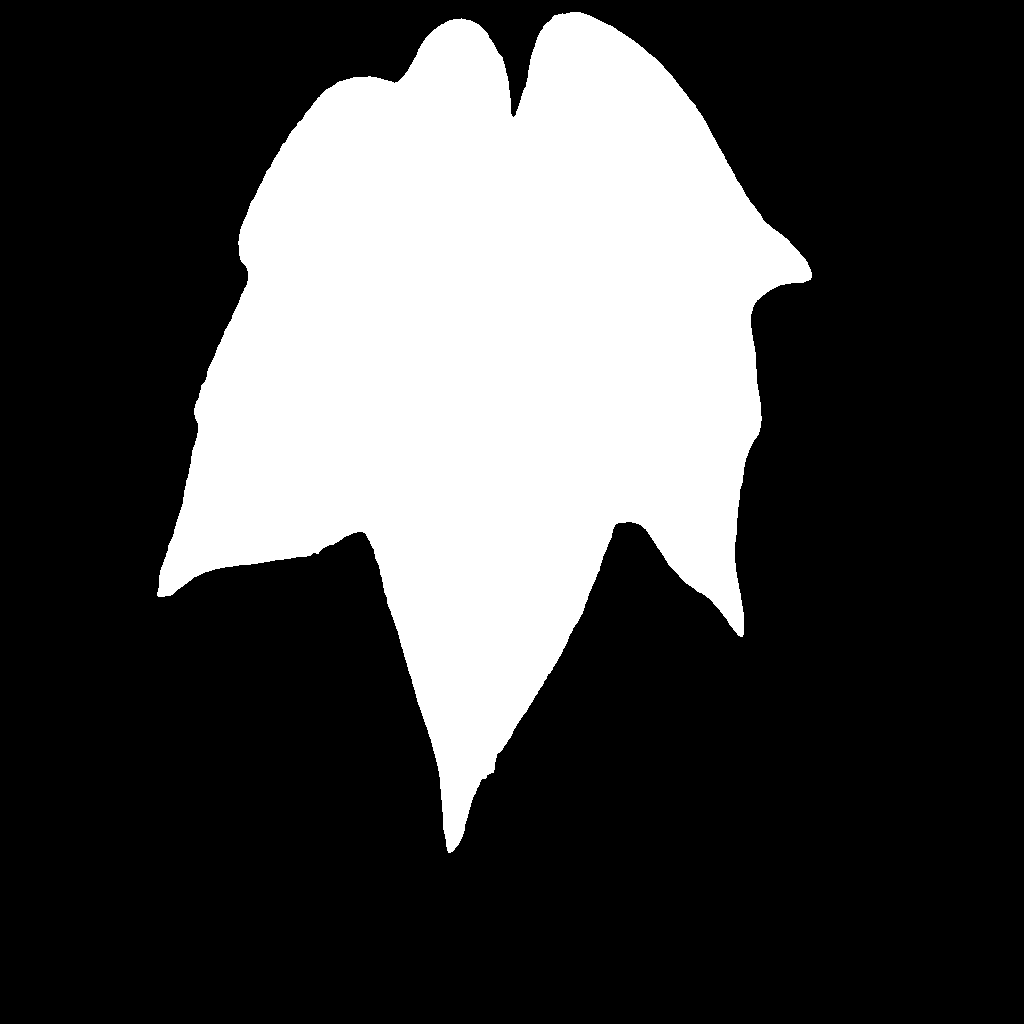}
    \end{minipage} 
    \begin{minipage}{0.24\linewidth}
         \centering
             \includegraphics[width=0.99\linewidth]{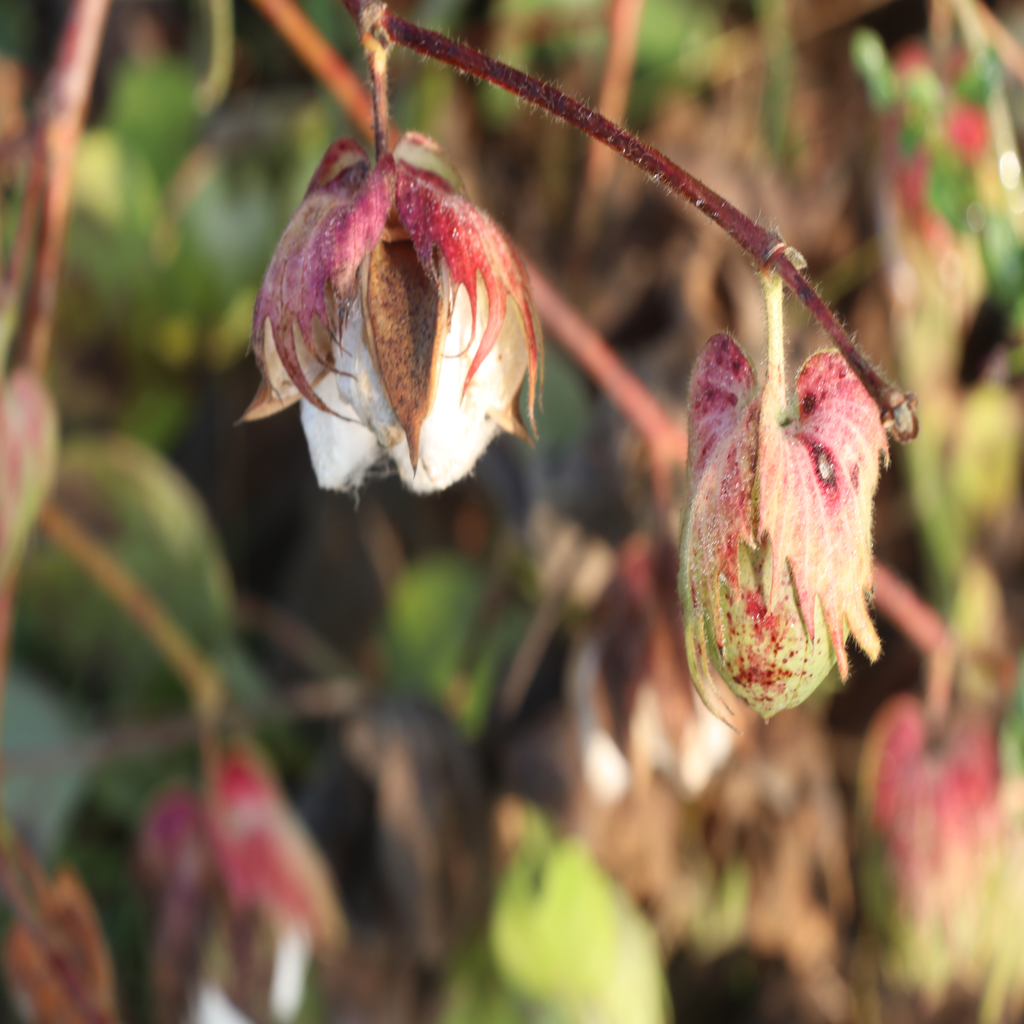}
    \end{minipage}       
    \begin{minipage}{0.24\linewidth}
         \centering
             \includegraphics[width=0.99\linewidth]{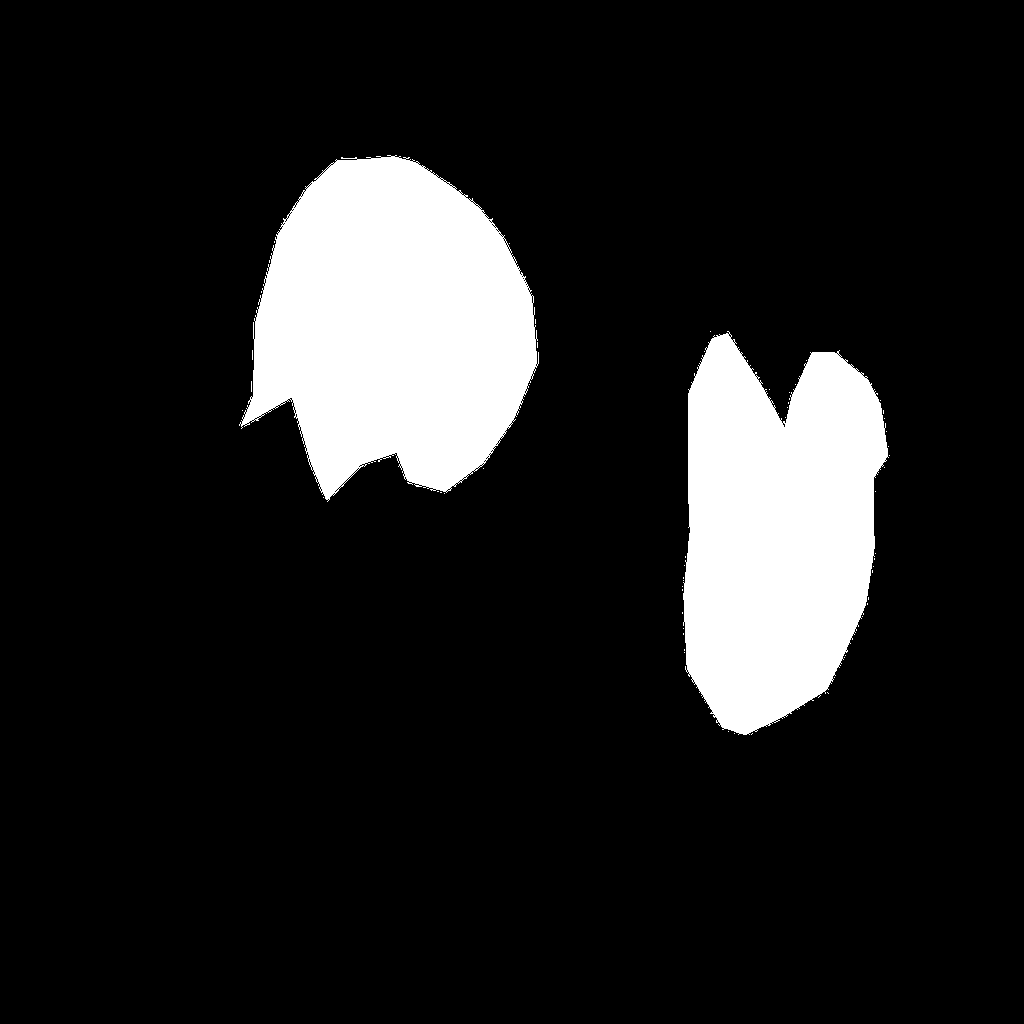}
        \end{minipage} 
    \begin{minipage}{0.24\linewidth}
         \centering
             \includegraphics[width=0.99\linewidth]{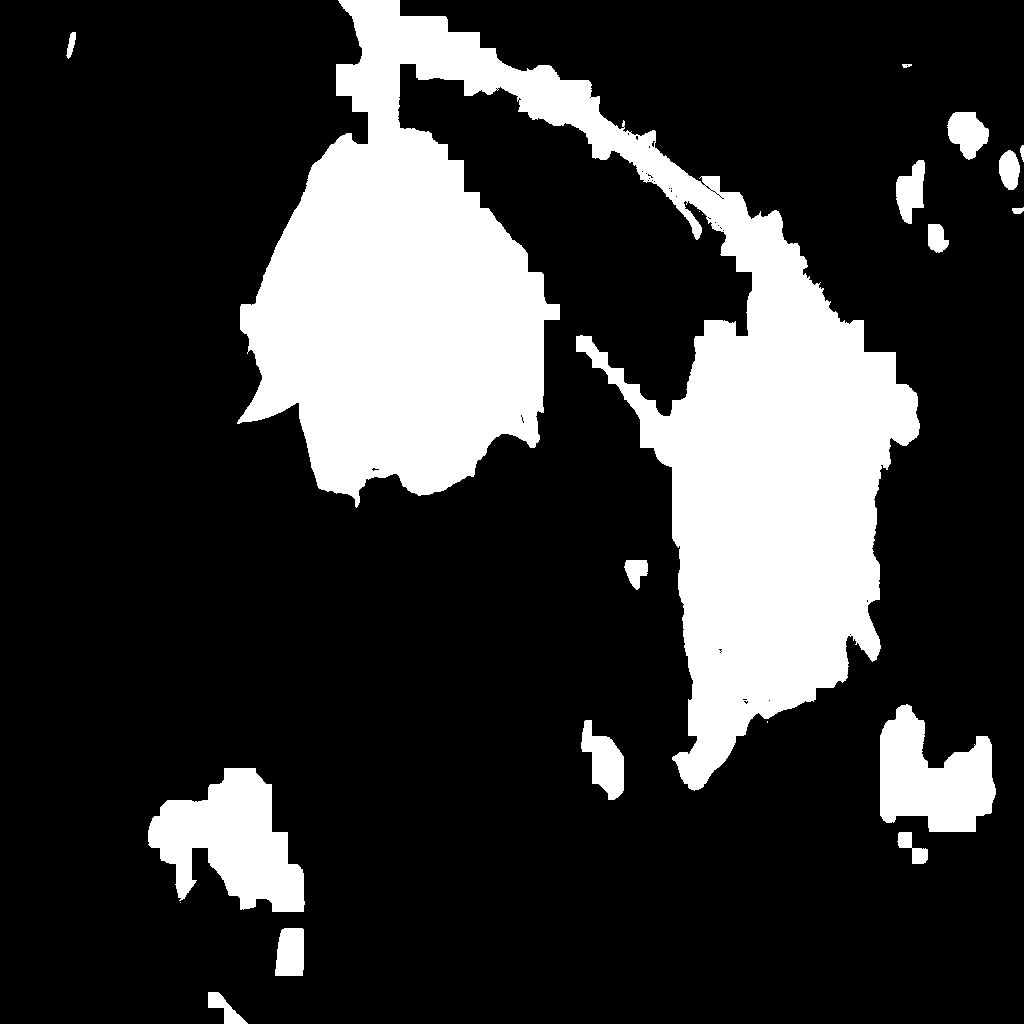}
    \end{minipage}    
   \begin{minipage}{0.24\linewidth}
         \centering
             \includegraphics[width=0.99\linewidth]{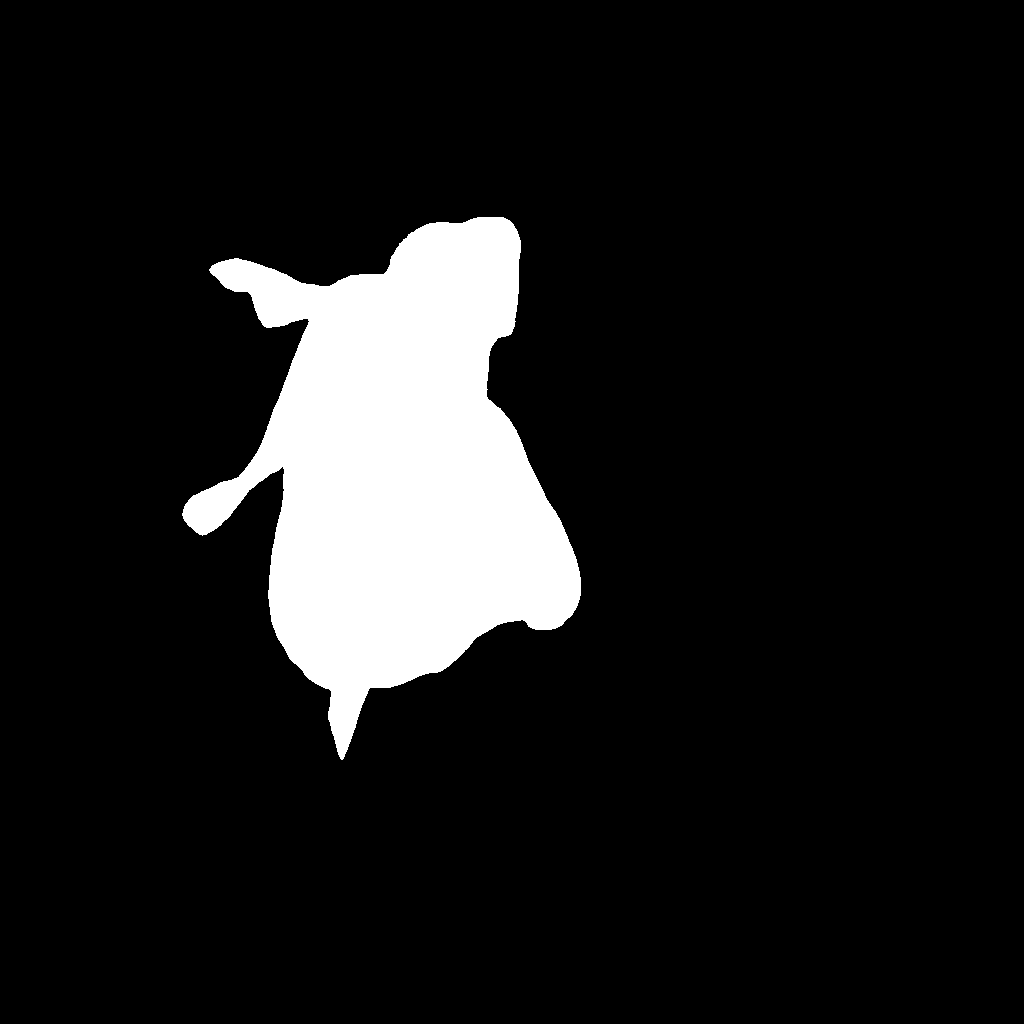}
    \end{minipage} 
    % \begin{minipage}{0.24\linewidth}
    %      \centering
    %          \includegraphics[width=0.99\linewidth]{images/Object_Segmentation/IMG_8975.png}
    % \end{minipage} 
    % \begin{minipage}{0.24\linewidth}
    %      \centering
    %          \includegraphics[width=0.99\linewidth]{images/Object_Segmentation/IMG_8975_bm.png}
    % \end{minipage}  
    % \begin{minipage}{0.24\linewidth}
    %      \centering
    %          \includegraphics[width=0.99\linewidth]{images/Object_Segmentation/IMG_8975_DSM.png}
    % \end{minipage}
    % \begin{minipage}{0.24\linewidth}
    %      \centering
    %          \includegraphics[width=0.99\linewidth]{images/Object_Segmentation/IMG_8975_SAM.png}
    % \end{minipage}      
    \caption{Object Segmentation with supervised (SAM) and unsupervised~\cite{melas2022deep} methods.}
    \label{fig:Obj_Seg}   
\end{figure}

\subsection{Object Segmentation and Restoration}
 we used  Segment Anything Model (SAM), and  the Deep Spectral method\cite{melas2022deep}. DSM extracts features from a self-supervised pre-trained network and uses spectral graph theory on feature correlations to obtain eigenvectors. These eigensegments correspond to semantically meaningful regions with well-defined boundaries. 

% Qualitative outputs on the proposed dataset are shown in Fig(~\ref{fig:Obj_Seg}).

Image restoration is essential for addressing noise, artefacts, low resolution, and missing data in computer vision. We explored a robust unsupervised restoration~\cite{poirier2023robust}  framework using a StyleGAN-based architecture and the Bilateral Graph Regularization Model (BGRM) for improved restoration across degradation scenarios. Evaluated on denoising, deartifacting, upsampling, and inpainting tasks with drone(see Table~\ref{table:RUSIR_mul}) and DSLR image datasets.

% \begin{figure}[!h]
%   \includegraphics[width=\linewidth]{images/restore.png}
%   \caption{Image Restoration Results .}
%   \label{fig:pred_comp_det}
% \end{figure}

% \begin{figure}[!h]
%   \includegraphics[width=\linewidth]{images/udta.png}
%   \caption{Folder structure of the COT-AD Dataset Annotated. The directory, \textit{Aerial Images for Detection and Segmentation}, is divided into four parts over six months, each containing subfolders for images, detection labels, segmentation masks, and segmentation labels. The \textit{High-Resolution DSLR Images for Cotton Crop Insights} directory is organized into categories of Leaf disease, Cotton Boll, and Bugs, each with specific subdirectories for detailed analysis.}
%   \label{fig:folder_struc}
% \end{figure}

\begin{table*}[!t]
\centering
\renewcommand{\arraystretch}{1.3} % Improve row spacing
\resizebox{0.9\textwidth}{!}{
\begin{tabular}{ l c c }
\hline
\textbf{Model} & \textbf{\begin{tabular}[c]{@{}l@{}}Zero Shot\\  (Top-5 )\end{tabular}} & \textbf{\begin{tabular}[c]{@{}l@{}}Linear Probing\\ (Accuracy)\end{tabular}} \\
\hline
\multicolumn{3}{c}{\textbf{Augmented (Cotton Leaf Disease Detection) Dataset\cite{bishshash2024comprehensive}}} \\   
\hline
CLIP (ViT-B/32)~\cite{radford2021learning} & 93.67 & 91.54 \\
CLIP (ViT-B/16)~\cite{radford2021learning} & 96.72 & \textbf{95.86} \\
CLIP (RN50)~\cite{radford2021learning}     & \textbf{98.88} & 88.74 \\
BioCLIP~\cite{stevens2024bioclip}          & 67.41 & 94.54 \\
BioCLIP (ViT-B/16, iNat-only)~\cite{stevens2024bioclip} & 69.98 & 93.21 \\
\hline
\multicolumn{3}{c}{\textbf{Original (Cotton Leaf Disease Detection) Dataset\cite{bishshash2024comprehensive}}} \\
\hline
CLIP (ViT-B/32)~\cite{radford2021learning} & 91.76 & 90.74 \\
CLIP (ViT-B/16)~\cite{radford2021learning} & 92.85 & 93.38 \\
CLIP (RN50)~\cite{radford2021learning}     & \textbf{95.73} & 84.37 \\
BioCLIP~\cite{stevens2024bioclip}          & 61.98 & \textbf{91.87} \\
BioCLIP (ViT-B/16, iNat-only)~\cite{stevens2024bioclip} & 64.93 & 91.36 \\
\hline
\multicolumn{3}{c}{\textbf{cotton leaf disease dataset\cite{cotton_leaf_disease_dataset}}} \\
\hline
CLIP (ViT-B/32)~\cite{radford2021learning} & 92.40 & 87.85 \\
CLIP (ViT-B/16)~\cite{radford2021learning} & 94.95 & \textbf{90.37} \\
CLIP (RN50)~\cite{radford2021learning}     & \textbf{96.83} & 84.96 \\
BioCLIP~\cite{stevens2024bioclip}          & 64.81 & 88.36 \\
BioCLIP (ViT-B/16, iNat-only)~\cite{stevens2024bioclip} & 66.32 & 89.76 \\
\hline
\multicolumn{3}{c}{\textbf{A Dataset of Cotton Leaf Images for Disease Detection and Classification\cite{miranidataset}}} \\
\hline
CLIP (ViT-B/32)~\cite{radford2021learning} & 94.00 & 86.69 \\
CLIP (ViT-B/16)~\cite{radford2021learning} & 95.88 & 89.43 \\
CLIP (RN50)~\cite{radford2021learning}     & \textbf{97.62} & 84.86 \\
BioCLIP~\cite{stevens2024bioclip}          & 65.27 & 87.46 \\
BioCLIP (ViT-B/16, iNat-only)~\cite{stevens2024bioclip} & 67.66 & \textbf{90.96} \\
\hline
\end{tabular}}
\caption{Image Classification: comparison of models on Zero-Shot (Top-5) and Linear Probing (Top-1) accuracy across four datasets.}
\label{tab:merged_top5_top11}
\end{table*}

\section{Data Acquisition and Annotation}
The acquisition of data for our crop detection and segmentation dataset was meticulously planned to capture the detailed growth cycles of cotton crops. Using drone technology, we conducted aerial surveys of two cotton fields throughout their life cycle. The drones operated at altitudes [10,15,115] meters, providing a balanced perspective for capturing high-resolution images without compromising the detail necessary for precise segmentation and detection tasks. Drone flights were scheduled at regular intervals to ensure that each phase of crop growth was adequately documented. This systematic capture process allowed us to monitor the development stages of cotton plants. In addition, it facilitated the detection of different agricultural phenomena, including the spread of weeds, pest infestations, and disease outbreaks.

\begin{figure}[!t]
  \centering
  \includegraphics[width=0.9\linewidth]{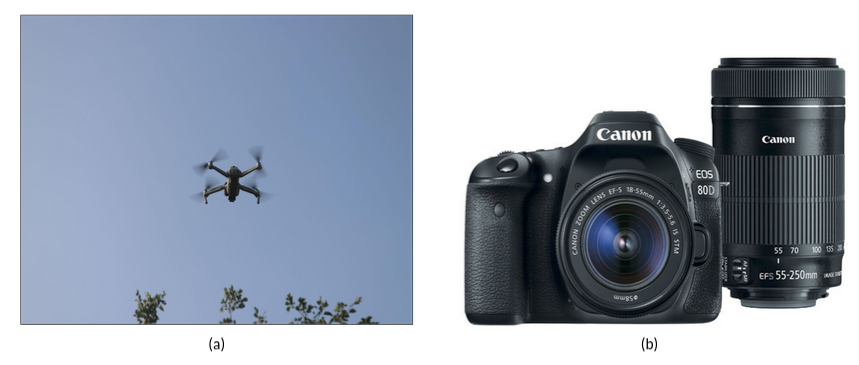}
  \caption{Data Acquisition Equipment: (a) DJI MAVIC Air 2s (b) DSLR Camera Canon 80D (w) \textit{Image Source: Canon Asia)}}
  \label{fig:drone_camera}
\end{figure}

\section{Hardware Specification for Data Collection}
\label{appendix:A}
\textbf{RGB Drone (DJI MAVIC Air 2s) ---} DJI Mavic Air 2s capture Red, Green, and Blue- visible spectra, creating humanly recognizable coloured images with a maximum flight distance of up to 18.5 km and with a maximum flight time of around 30 minutes (under no wind conditions). It has GNSS connectivity of GPS + GLONASS + Galileo constellations. A camera with 20 MP is integrated with the drone on a 3-axis Gimbal. Figure \ref{fig:drone_camera} shows the DJI Mavic Air 2s flying for the survey.

\textbf{DSLR Camera (Canon EOS 80D (W)) ---} Canon EOS 80D (W) is a versatile and high-performance digital single-lens reflex camera with a CMOS sensor with approximately 24.2 megapixels and DIGIC 6 image processor. The 45-point all-cross-type autofocus system allows high precision and high-speed performance with a maximum of approximately 7.0 fps continuous shooting and wireless functions. Figure~\ref{fig:drone_camera} shows the Canon EOS 80D (W) used in capturing cotton disease images.

\section{Insights from the Fields}
\label{appendix:B}

Table \ref{table:month_wise} provides a detailed month-by-month analysis of cotton farms using aerial imagery. This analysis helps understand the dataset in greater depth as it curates the crops across its growth stages throughout the season.

Table \ref{table:month_wise_disease} provides a detailed month-by-month analysis of diseases seen in the cotton crops. This analysis helps identify disease occurrence and progression patterns, denoting timely and effective data collection and stats throughout the growing season.

\section{Dataset for Disease Detection in Crops}
Cotton Crops are susceptible to pests, diseases, climate changes, and weed competition. Primarily, the types of insects found during cultivation include Jassids/Aphids, Thrips, Whiteflies, and bollworms, and the diseases include bacterial blight, fungal leafspots, root rot, leaf curling, leaf reddening, etc. To support the management of diseases and pests, high-resolution images of the diseased part of the cotton plants were systematically captured from the cotton farm. Leaf diseases such as bacterial blight, leaf spots, leaf reddening, and pink boll rot were prominently present towards the end of the crop's lifecycle. Pesticides were intentionally avoided on the farm to allow diseases to spread naturally. Experts meticulously captured the abnormalities present in the plants to develop a comprehensive dataset. Red Cotton Bugs and Mealy Bugs were notably present in certain farm sections during their maturity period.

The diseases and bug infestations were the primary reason to capture images with a high-resolution DSLR camera. The collection of the \textit{High-Resolution DSLR Images of Cotton Crop Insights} aims to provide a benchmark for classifying cotton diseases(see Fig.~\ref{fig:seg_performancebghg}). It can be used to get information about the diseases present on the farm, such as their names, causes, measures to be taken, and pesticide information to decrease their effect. 

% \begin{figure}
%   \includegraphics[width=\linewidth]{images/img_det_training.png}
%   \caption{Graphical representation of training performance metrics for detection models YOLOv8 and DETR: (a) Curve showing the box loss of YOLOv8 over training epochs. (b) A curve showing mAP@50 (mean Average Precision at 50\% IoU threshold) for YOLOv8. (c) A curve showing the box loss of DETR through the training epochs. (d) Curve showing mAP@50 for DETR}
%   \label{fig:det_performance}
% \end{figure}

% \begin{figure}
%   \includegraphics[width=\linewidth]{images/img_bb_mask.png}
%   \caption{Overview of Detection and Segmentation Dataset - original aerial images of cotton crops, the corresponding detection bounding boxes, and the segmentation masks.}
%   \label{fig:bb_mask_overview}
% \end{figure}

\section{Dataset Access and Storage}
We have shared the COT-AD dataset on Kaggle as well as IEEE DataPort, which comprises approximately 308 GB of data.  Researchers and practitioners interested in accessing the dataset can visit our project page(https://aamaanakbar.github.io/COT-AD/), or we have also provided the link to the dataset in the Dataset documentation. The dataset is intended for non-commercial use.

\section{Dataset Organization}
\label{sec:data_org}

\textbf{Aerial Images for Detection and Segmentation:}
The data for detection and segmentation tasks is organized under the directory named \textit{Aerial Images for Detection and Segmentation}, which is segmented into four main parts:

\begin{itemize}
    \item \textbf{Part A:} Data from the first two months.
    \item \textbf{Part B:} Data from the third month.
    \item \textbf{Part C:} Data from the fourth month.
    \item \textbf{Part D:} Data from the fifth and sixth months.
\end{itemize}

Each part includes four subfolders:
\begin{itemize}
    \item \textbf{Images:} Holds \textit{JPG} aerial images of cotton crops.
    \item \textbf{Detection Labels:} Stores YOLO-formatted \textit{.txt} files for single-class detection tasks.
    \item \textbf{Segmentation Masks:} Holds \textit{JPG} binary masks for cotton crop segmentation.
    \item \textbf{Segmentation Labels:} Contains YOLO-formatted \textit{.txt} files tailored for segmentation tasks.
\end{itemize}

\textbf{Dataset for Disease Classification:}
Organized under the directory \textit{High-Resolution DSLR Images for Cotton Crop Insights}, the classification of images is based on three categories -- Leaf Disease, Cotton Boll and Bugs.

\begin{itemize}
    \item The \textbf{Leaf} folder contains the four subdirectories -- Yellowish Leaf, Leaf Spot, Bacterial Blight, Leaf Reddening, and Fresh Leaf. 
    \item The \textbf{Cotton Boll} folder contains the three subdirectories -- Boll Rot, Damaged Cotton Boll, and Healthy Cotton Boll.
    \item The \textbf{Bugs} folder contains two subdirectories -- Mealy Bug and Red Cotton Bug.
\end{itemize}

This organization enhances the accessibility and efficiency of data processing for various model training purposes related to detection, segmentation and disease classification tasks.

\section{Results}
Image classification on different data presented in table(1) in the main paper, we have done classification and segmentation tasks on that data. See the result in Table ~\ref{tab:merged_top5_top11}
. We performed image enhancement using multiple datasets and a variety of methods. The qualitative results of these enhancement experiments are illustrated in Figure~\ref{fig:img_enhan_datasets}.

% \begin{table}[!h]
% \centering
% \resizebox{0.49\textwidth}{!}{ % Adjust width to fit half-page
% \begin{tabular}{ c c c }
% \hline
% \textbf{Model} & \textbf{\begin{tabular}[c]{@{}l@{}}Zero Shot\\  (Top-5 )\end{tabular}} & \textbf{\begin{tabular}[c]{@{}l@{}}Linear Probing\\ (Accuracy)\end{tabular}} \\
% \hline
% CLIP (ViT-32)~\cite{radford2021learning} & 93.99 & 89.79 \\
% CLIP (ViT-16)~\cite{radford2021learning} & 95.88 & 92.27 \\
% CLIP (RN50)~\cite{radford2021learning} & \textbf{97.61} & 86.39 \\
% Imageomics/bioclip~\cite{stevens2024bioclip} & 65.27 & \textbf{91.96} \\
% Bioclip-vit-b-16-inat-only~\cite{stevens2024bioclip} & 67.65 & 91.49 \\
% \hline
% \end{tabular}}
% \caption{Model Performance: Zero-Shot and Linear Probing}
% \label{tab:zeroshot_linearProbing}
% \end{table}

\begin{figure*}[!h]
    \centering

    % Row 1
    \begin{minipage}{0.9\linewidth}
        \centering
        \includegraphics[width=0.92\linewidth]{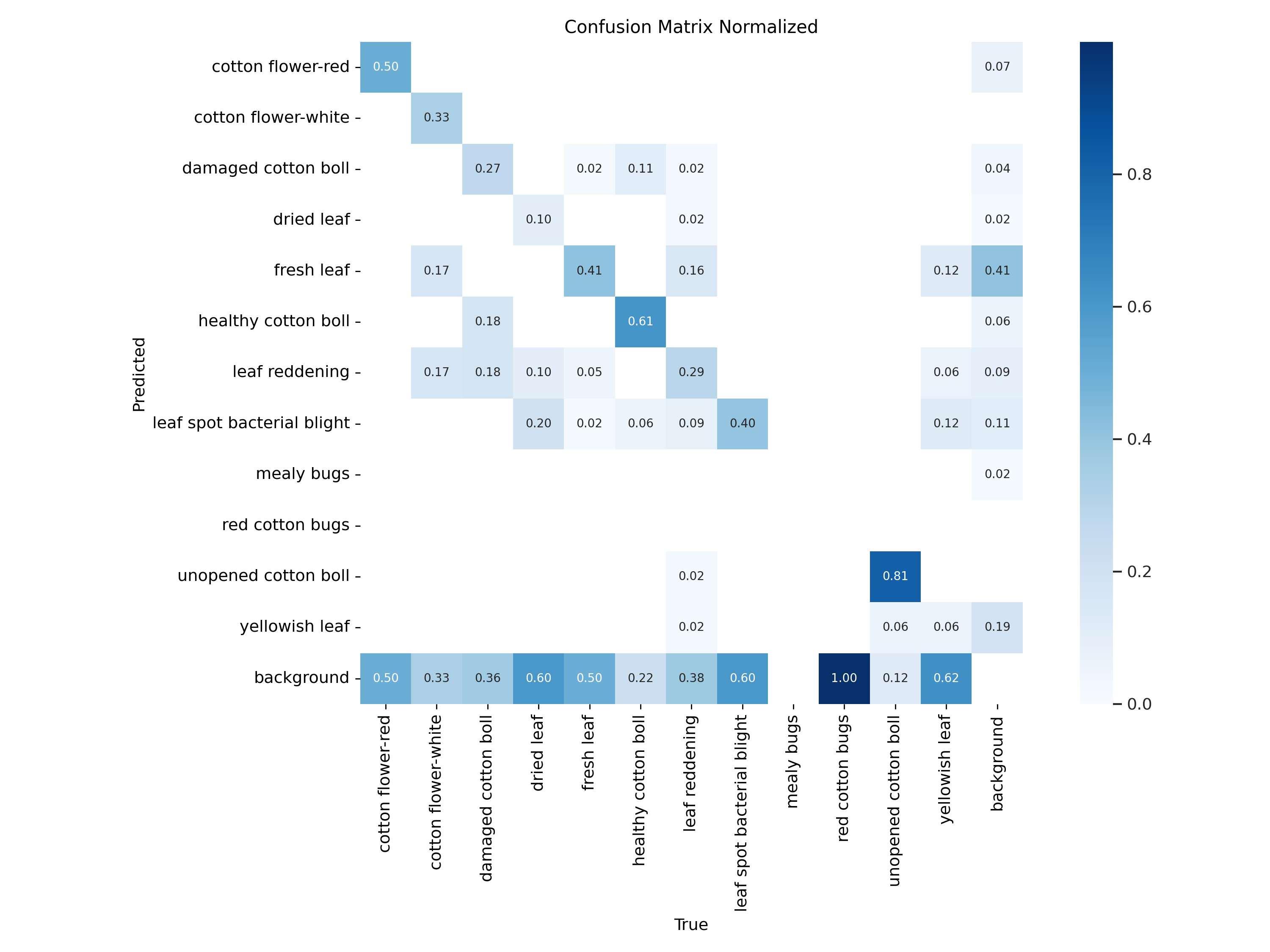}
        \caption*{(a) \textbf{large} — \texttt{normalised confusion matrix}}
    \end{minipage}
    % \hfill
    % \begin{minipage}{0.45\linewidth}
    %     \centering
    %     \includegraphics[width=0.92\linewidth]{images/yolov11/dronedata/l_m.jpg}
    %     \caption*{(b) \textbf{Medium} — \texttt{s_m.jpg}}
    % \end{minipage}

    \vspace{1em}

    % Row 2
    \begin{minipage}{0.9\linewidth}
        \centering
        \includegraphics[width=0.92\linewidth]{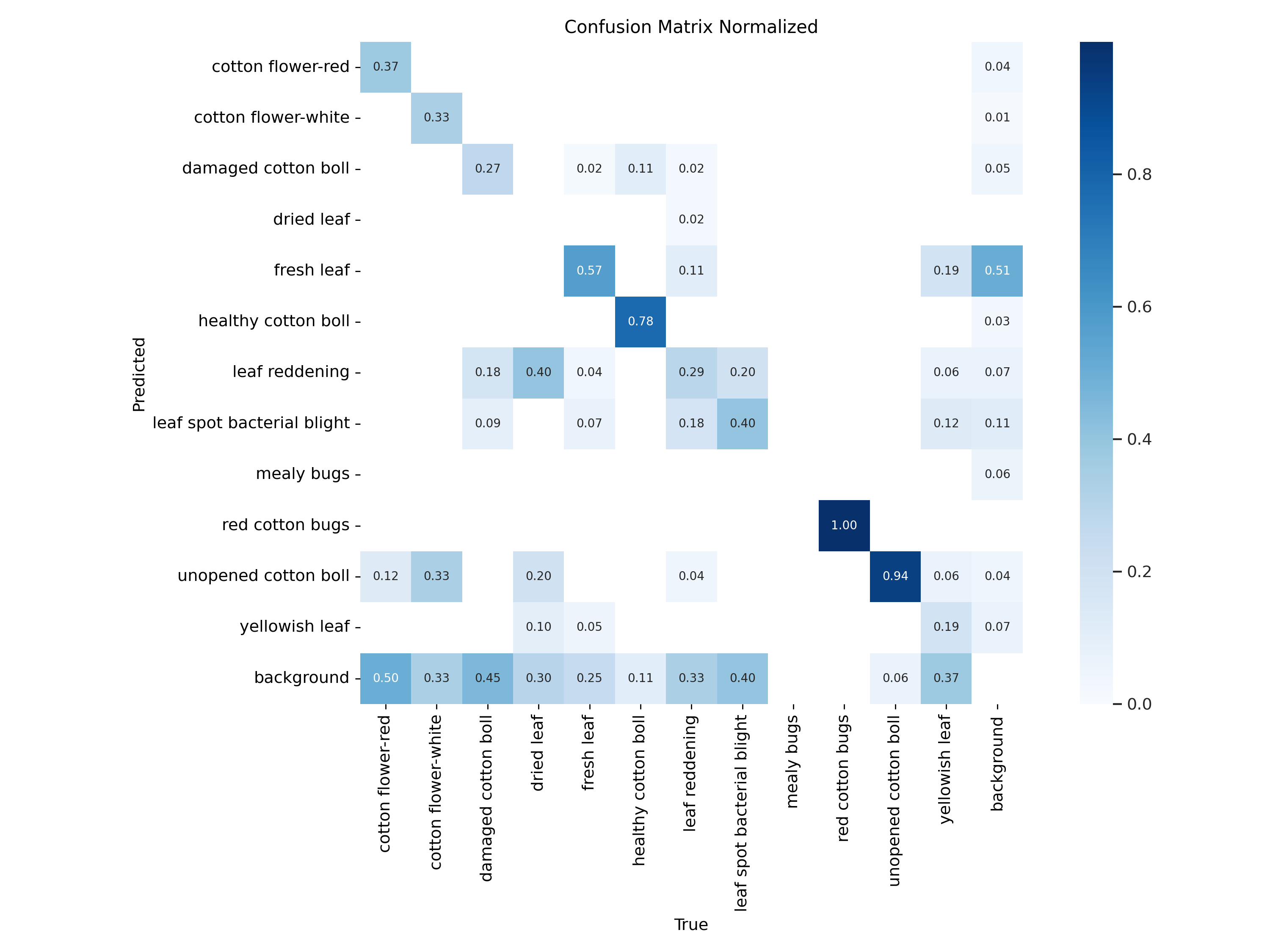}
        \caption*{(c) \textbf{Xtra} — \texttt{Normalised Confusion Matrix}}
    \end{minipage}
    % \hfill
    % \begin{minipage}{0.45\linewidth}
    %     \centering
    %     \includegraphics[width=0.92\linewidth]{images/yolov11/dronedata/val_batch0_pred.jpg}
    %     \caption*{(d) \textbf{Xtra} — \texttt{img00000018\_AesPA-Net.png}}
    % \end{minipage}

    \caption{\textbf{ Image Predictions.} Displaying predictions Score on DSLR data across different categories:  Large and Xtra. }
    \label{fig:yolov11_preds_named}
\end{figure*}

\begin{figure*}[!h]
    \centering

    % Row 1
    \begin{minipage}{0.9\linewidth}
        \centering
        \includegraphics[width=0.92\linewidth]{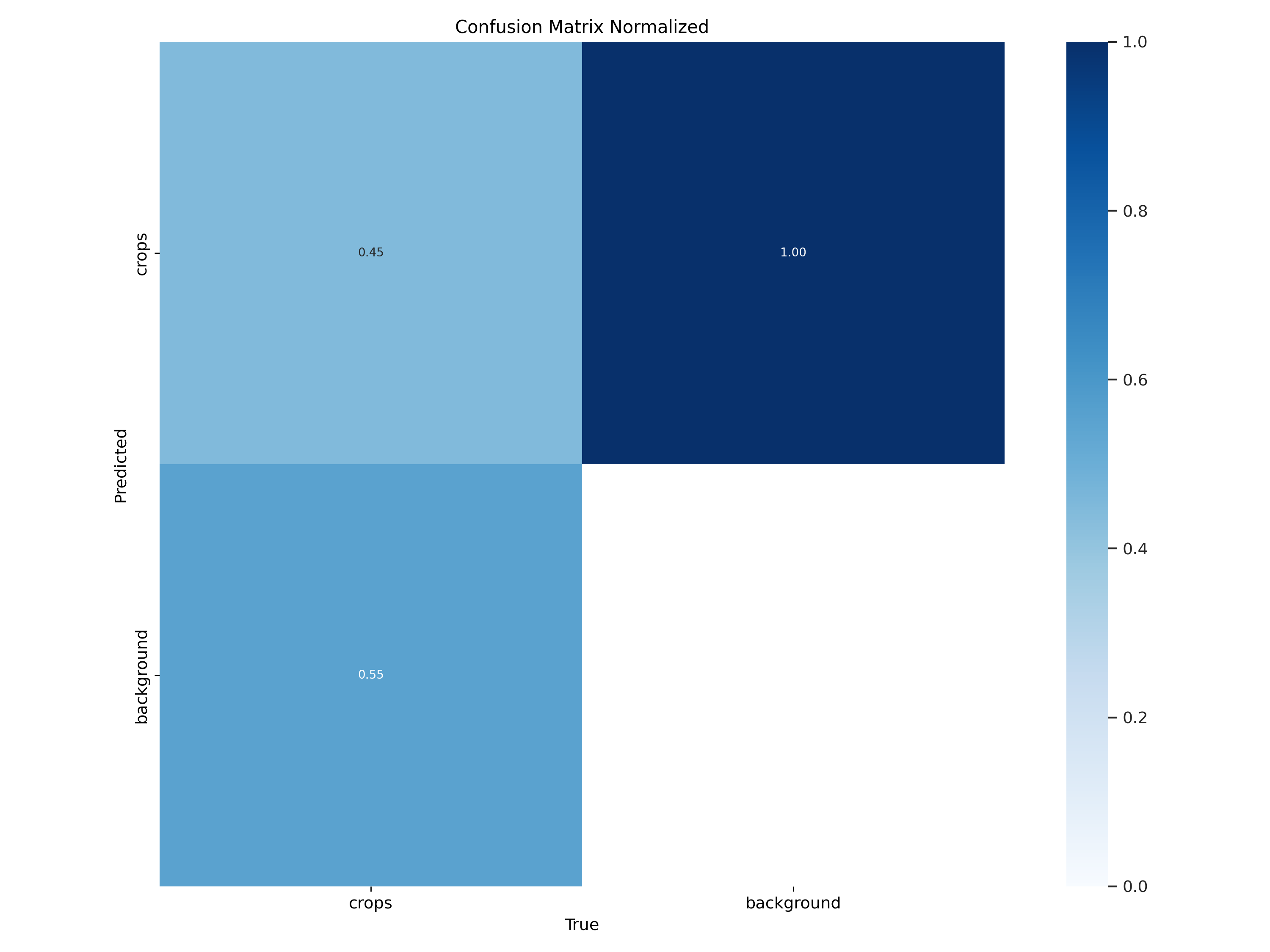}
        \caption*{(a) \textbf{large} — \texttt{normalised confusion matrix}}
    \end{minipage}
    % \hfill
    % \begin{minipage}{0.45\linewidth}
    %     \centering
    %     \includegraphics[width=0.92\linewidth]{images/yolov11/dronedata/l_m.jpg}
    %     \caption*{(b) \textbf{Medium} — \texttt{s_m.jpg}}
    % \end{minipage}

    \vspace{1em}

    % Row 2
    \begin{minipage}{0.9\linewidth}
        \centering
        \includegraphics[width=0.92\linewidth]{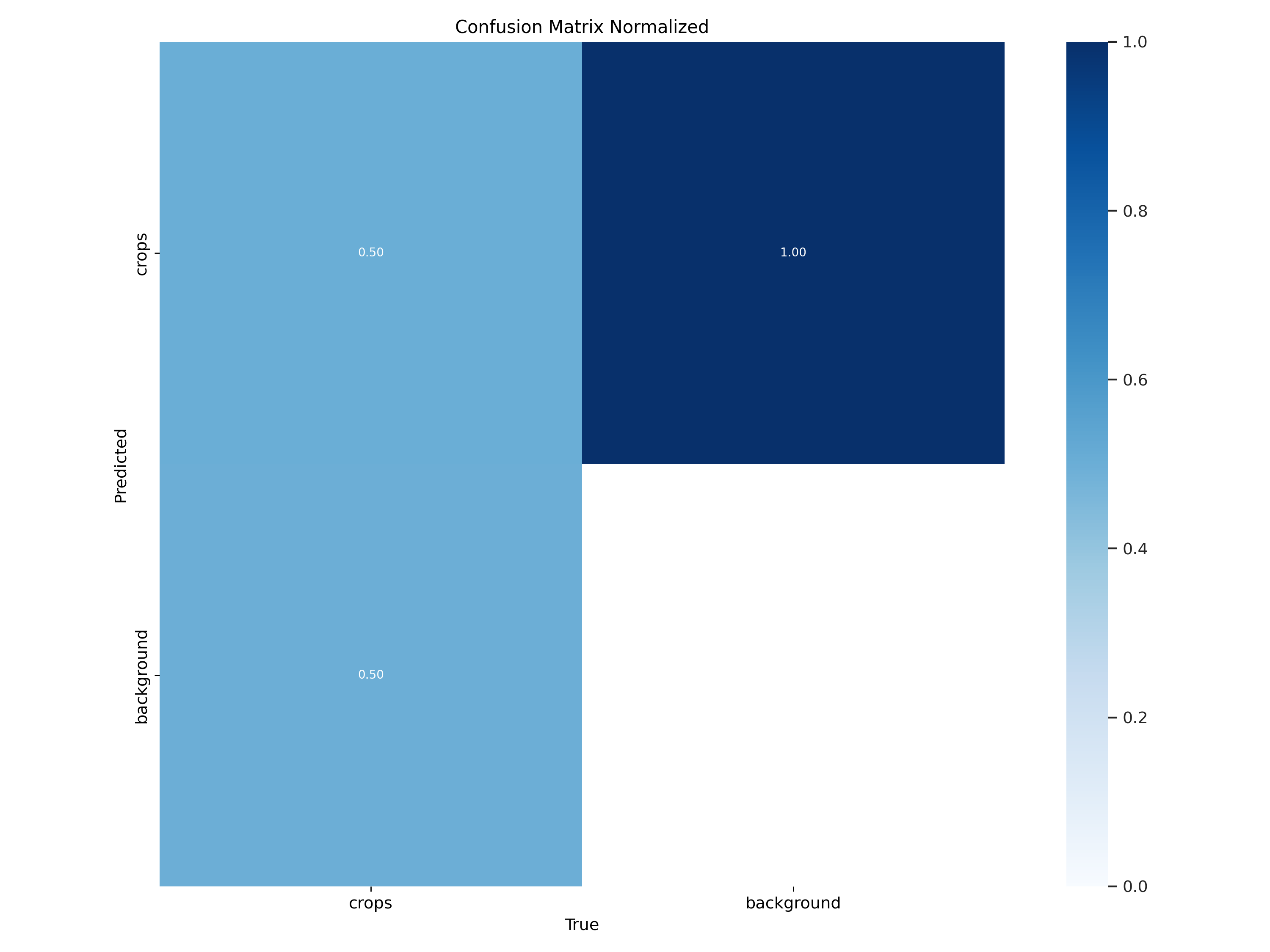}
        \caption*{(c) \textbf{Xtra} — \texttt{Normalised Confusion Matrix}}
    \end{minipage}
    % \hfill
    % \begin{minipage}{0.45\linewidth}
    %     \centering
    %     \includegraphics[width=0.92\linewidth]{images/yolov11/dronedata/val_batch0_pred.jpg}
    %     \caption*{(d) \textbf{Xtra} — \texttt{img00000018\_AesPA-Net.png}}
    % \end{minipage}

    \caption{\textbf{Image Predictions.} Displaying predictions Score on Drone data across different categories:  Large and Xtra. }
    \label{fig:yolov11_preds_named3}
\end{figure*}

For image prediction, we employed the YOLOv11 model and presented the results using confusion matrices for both DSLR and drone datasets in the 
Fig(~\ref{fig:yolov11_preds_named},~\ref{fig:yolov11_preds_named3})

\section*{Acknowledgments}
This research and data collection were supported by \textbf{ L\&T Technology Services Limited}, Vadodara, India, in collaboration with the Indian Institute of Technology, Gandhinagar, India and Jibaben Patel Chair in AI. We thank the research group of L\& T Technology Services, specifically Mr Ashokkumar Jain, Mr Pushkar Shaktawat, and Mr Nikhil Dev.  Additionally, we would like to acknowledge the efforts of  Mr Prakram Singh Rathore, Mr Medhansh Singh, Mr Pavidhar Jain, Mr Yash Khandelwal, and Mr Harshvardhan Vala at the Indian Institute of Technology Gandhinagar for help with annotating the aerial images.

% \bibliographystyle{IEEEbib}
% \bibliography{strings}

\end{document}